\documentclass[10pt,twocolumn,letterpaper]{article}

\usepackage{wacv}
\usepackage{times}
\usepackage{epsfig}
\usepackage{graphicx}
\usepackage{amsmath}
\usepackage{amssymb}
\usepackage[accsupp]{axessibility}

\wacvfinalcopy % *** Uncomment this line for the final submission

\usepackage[pagebackref=true,breaklinks=true,colorlinks,bookmarks=false]{hyperref}
\begin{document}

%%%%%%%%% TITLE
\title{Self-Supervised Knowledge Transfer via Loosely Supervised Auxiliary Tasks}

\author{Seungbum Hong, Jihun Yoon, and Min-Kook Choi\\
VisionAI, hutom\\
Seoul, Republic of Korea\\
{\tt\small \{qbration21, jhyoon2020, mkchoi\}@hutom.io}
\and
Junmo Kim\\
KAIST\\
Daejeon, Republic of Korea\\
{\tt\small junmo.kim@kaist.ac.kr}
}

\maketitle
\thispagestyle{empty}

%%%%%%%%% ABSTRACT
\begin{abstract}
Knowledge transfer using convolutional neural networks (CNNs) can help efficiently train a CNN with fewer parameters or maximize the generalization performance under limited supervision. To enable a more efficient transfer of pretrained knowledge under relaxed conditions, we propose a simple yet powerful knowledge transfer methodology without any restrictions regarding the network structure or dataset used, namely self-supervised knowledge transfer (SSKT), via loosely supervised auxiliary tasks. For this, we devise a training methodology that transfers previously learned knowledge to the current training process as an auxiliary task for the target task through self-supervision using a soft label. The SSKT is independent of the network structure and dataset, and is trained differently from existing knowledge transfer methods; hence, it has an advantage in that the prior knowledge acquired from various tasks can be naturally transferred during the training process to the target task. Furthermore, it can improve the generalization performance on most datasets through the proposed knowledge transfer between different problem domains from multiple source networks. SSKT outperforms the other transfer learning methods (KD, DML, and MAXL) through experiments under various knowledge transfer settings. The source code will be made available to the public\footnote{\url{https://github.com/generation21/generation6011}}.
\end{abstract}

%%%%%%%%% BODY TEXT
\section{Introduction}

Knowledge transfer is the most representative training methodology for improving the generalization capability and training efficiency of convolutional neural networks (CNNs). The most widely used knowledge transfer is transfer learning \cite{Yosinski14,Tan18}, which uses pretrained weights trained on large-scale datasets as initial values for new tasks. Pretrained weights have been used as feature encoders after fine-tuning in different vision tasks such as image classification, object detection, and semantic segmentation \cite{Liu16,Long15,Ren15}. Taskonomy \cite{Zamir18,Zamir20} provides an extensive database and analysis approach for studying the effects of transfer learning using pretrained networks on the fine-tuning of target tasks. However, transfer learning using pretrained networks presupposes structural dependencies that must share essentially the same (whole or partial) network structure as the source network. Moreover, questions have been raised on whether transfer learning using pretrained weights can provide appropriate initial weights, which can help in improving convergence or training with less amount of data but not very effective for knowledge transfer to different tasks \cite{He19}.

\begin{figure}[t!]
\centering
\includegraphics[width=1.0\linewidth]{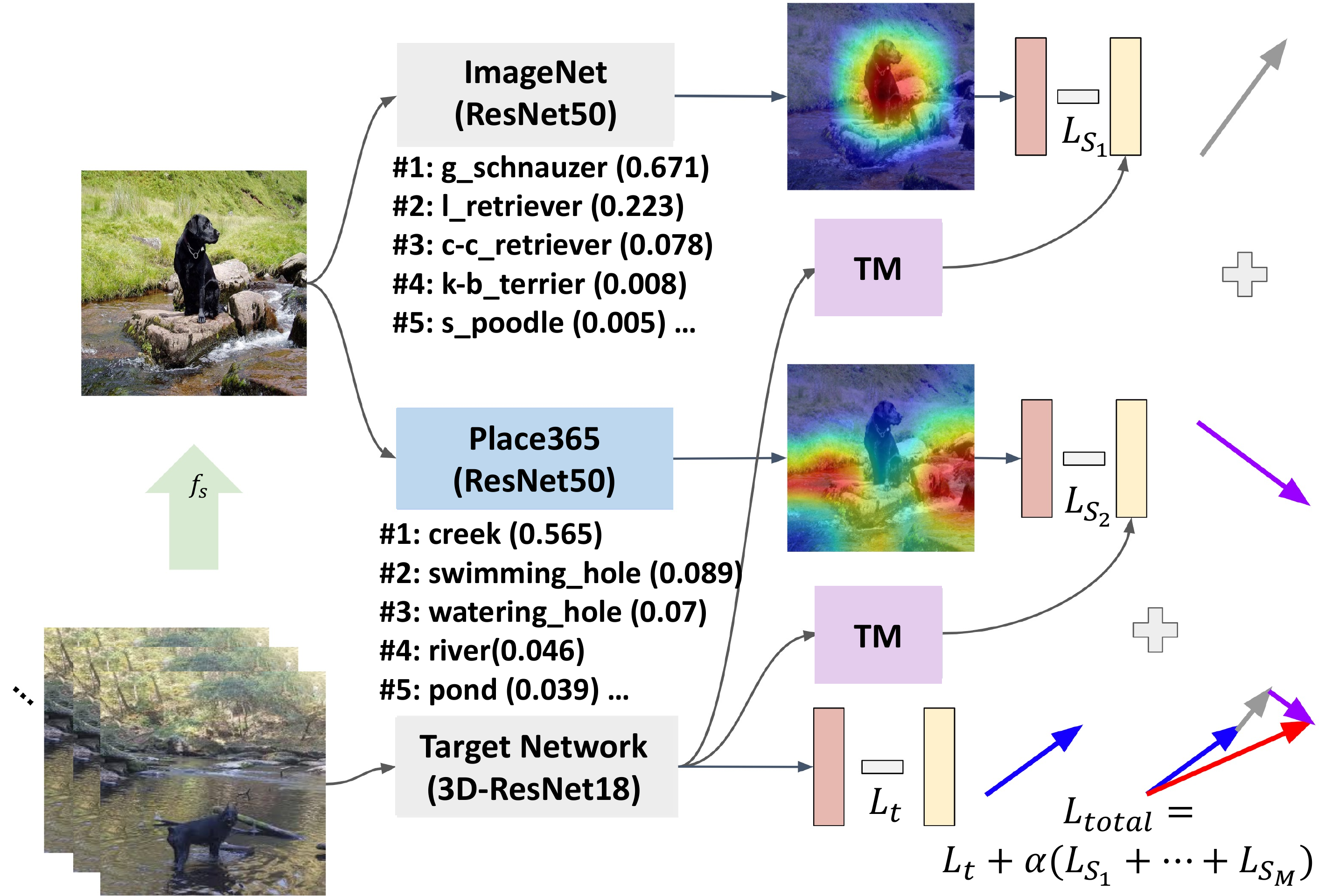}
\vspace{-3mm}
\caption{\textbf{Motivation of the SSKT.} The CNN responses that can be obtained from the same image vary depending on the type of supervision. SSKT performs auxiliary training using soft labels to convey the previously trained prior knowledge from the source tasks. At this time, the supervision used in the auxiliary training generates a gradient (grey and purple arrows) that can improve the generalization performance of the target task. The gradient to update the weight of the target network is obtained as a linear combination of all losses (red arrow). TM is a transfer module.}
\label{fig:example}
\vspace{-6mm}
\end{figure}

Another representative example of knowledge transfer using CNN is knowledge distillation (KD) \cite{Hinton14}, where pretrained teacher networks distill and deliver dark (hidden) knowledge in the inference output or learned features during student network training. KD methods include loss-based KD, which delivers dark knowledge through loss with a soft label \cite{Hinton14}, and KD, wherein the similarity between the features extracted are exploited in specific stages of the CNN \cite{Romero15}. Knowledge transfer through KD has the advantage of training student networks that have a better-generalized performance with fewer parameters comparable to teacher networks. In the general KD setup, the training datasets of both teacher and student networks should be the same. However, for a more general knowledge transfer, these constraints must be relaxed.

\begin{figure*}[t!]
\centering
\includegraphics[width=1.0\linewidth]{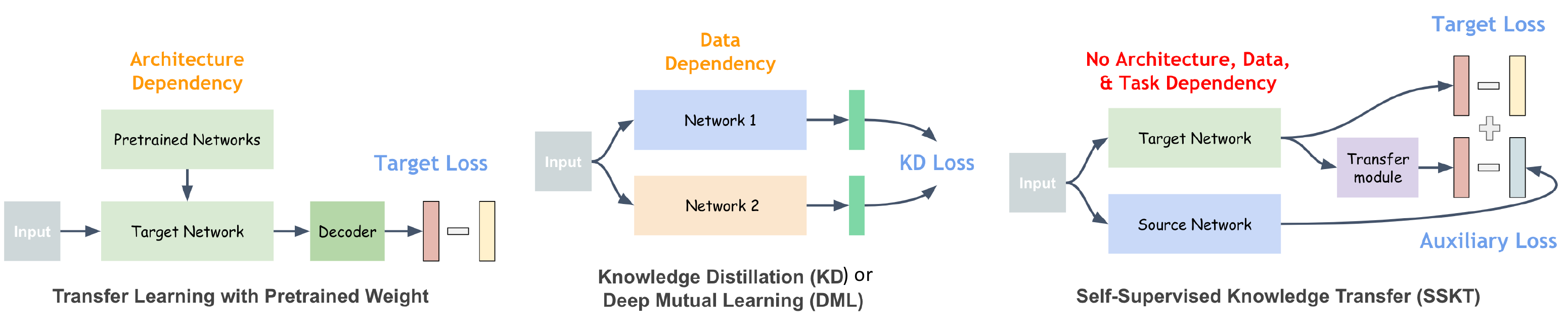}
\vspace{-5mm}
\caption{\textbf{Schematic of the difference between the proposed SSKT and other knowledge transfer methods.} Unlike existing knowledge transfer methods, such as KD and DML, SSKT is independent of the network structure and dataset. In addition, generalization performance improvements can be achieved without additional supervision during the training process.}
\label{fig:example}
\vspace{-5mm}
\end{figure*}

Figure 1 shows an example in which the information to be transferred can vary depending on the type of supervision of each pretrained CNN through Grad-CAM visualization \cite{Ramprasaat17} for the same input image. For this purpose, we propose a simple yet powerful method that aims for better generalization with minimal additional supervision. The proposed knowledge transfer conveys prior knowledge as a soft label from multiple pretrained networks obtained using different types of supervision methods in a self-supervised manner. The proposed knowledge transfer method utilizes a pretrained network but is independent of the structure of the target network and the training dataset of the pretrained network to train the target task. The target network is updated through the linear combination of the target loss and multiple auxiliary losses with soft label-based self-supervision from the pretrained networks by trying to guess the type of knowledge inferred from the source network (Figure 1). We expect that the target network reflects dark knowledge as a gradient from multiple supervisions of existing training data through soft label-based self-supervision. Because it does not refer to previously trained data and performs auxiliary training in a self-supervised manner, the proposed knowledge transfer can be performed within any task while requiring only the final output dimension (number of classes) for calculating each auxiliary loss. We call this form of knowledge transfer \textit{self-supervised knowledge transfer} (SSTK). The technical contributions of the SSKT are as follows:

\vspace{-3mm}

\begin{itemize}

\smallskip \item Unlike previous knowledge transfer techniques, SSKT does not depend on the structure of the source network or the training dataset. SSKT is even possible between structurally complete heterogeneous networks, such as 2D to 3D CNN knowledge transfer. \vspace{-1mm}

\smallskip \item SSKT has the potential for simultaneous knowledge transfer from multiple pretrained networks trained on completely different types of datasets or supervision. \vspace{-1mm}

\smallskip \item The SSKT improves the generalization performance in most knowledge transfer scenarios. Particularly in different problem domains, SSKT outperforms existing knowledge transfer methods. \vspace{-1mm}

\end{itemize}

Figure 2 shows the differences between SSKT and other knowledge transfer methods. For clarity, we take the problem domain as a superset of tasks (e.g., image classification, action recognition, object detection, and semantic segmentation), and the task is bounded in the same problem domain (e.g., ImageNet, Places365, and STL10 in image classification).

%------------------------------------------------------------------------

\section{Related Works}

\smallskip \noindent \textbf{Transfer learning with a pretrained initializer.} As the most commonly used method of knowledge transfer, transfer learning with a pretrained initializer has been applied to various fields of computer vision using high-performance models \cite{Krizhevsky12,Simonyan15,He16} with a large amount of training data. Because low-level feature information is shared by most visual recognition problems, fine-tuning is often applied by updating only a specific upper-layer when training new tasks \cite{Long15,Ren15}. Generally, this can be used in the same network structure, and many image recognition applications use only the feature encoder part of the entire task architecture \cite{Liu16,Long15,Ren15}. Although transfer learning using ImageNet has been successful so far, questions on transfer learning using ImageNet data have been raised \cite{He19}, and the experimental results of object recognition and key point recognition problems have been reported as examples where the pretrained initial values do not convey additional knowledge to the different problem domains. 

\noindent \textbf{Knowledge distillation.} Pretrained teacher networks with larger parameters can improve the generalization performance of student networks by performing knowledge transfer during the training process to a relatively small number of parameters \cite{Hinton14}. Student networks often show better generalization performance over teacher networks \cite{Furlanello18}. Recent knowledge distillation techniques for effective knowledge transfer used the relationship between the inferred posteriors from the teacher and student networks \cite{Park19} or the dataset as a subclass of the original classes to be learned \cite{Muller20}. In \cite{Furlanello18}, the ensemble model achieved a high recognition performance by distilling the knowledge of student networks with different initialization values using teacher and student networks with the same structure. SSKT uses a knowledge transfer method, such as KD with a soft label. However, unlike existing KD methods, because the SSKT transfers knowledge through an auxiliary task-based self-supervision technique, there is no dataset dependency. 

\noindent \textbf{Deep mutual learning (DML).} DML is one of the representative variations of the KD-based knowledge transfer method that uses multiple networks for efficient training through KD loss within networks \cite{Zhang18}. DML could achieve performance improvement through mutual learning in the middle of training without requiring a powerful pretrained teacher network. Recently, a learning methodology using feature fusion \cite{Kim20}, generative adversarial networks \cite{Chung20}, and collaborative learning \cite{Wu20} structures has been proposed to improve the efficiency of mutual learning during training. It differs from SSKT in that the DML performs multitask training-based learning on the same dataset and updates all the networks during training. In addition, it is fundamentally different from SSKT in that DML does not utilize prerequisite knowledge from former training. 

\noindent \textbf{Auxiliary learning.} Unlike general multi-task learning, in the case of auxiliary task learning, the purpose of a deep neural network is not to increase the performance of auxiliary tasks but to improve the performance of the target task. It was recently reported that auxiliary task learning could help improve the test performance of the target task, and the cosine similarity between the gradient losses of the target and auxiliary tasks was analyzed to visualize the cause of contribution by the auxiliary task \cite{Du19}. The basic structure of the SSKT was inspired by meta-learning techniques using auxiliary tasks under self-supervision (MAXL) \cite{Liu19}. However, MAXL requires a hierarchical auxiliary class structure defined manually for the primary class to be trained. SSKT was able to achieve higher performance than MAXL by using only the number of classes of source tasks as additional supervision information. 

\noindent \textbf{Self-supervised learning (SSL).} Typical self-supervised learning techniques train pretrained weights that perform unsupervised training on pretext tasks to obtain good initial weights \cite{Goyal19,Chen20_1,Chen20_2,He20}. The pretrained network utilized in SSKT was not obtained by unsupervised training, and the target task training differed from the general self-supervised approach, requiring meta information such as the number of classes used in the source network training. However, SSKT can still be considered self-supervised from the viewpoint of training auxiliary tasks using only soft labels from pretrained networks, without requiring additional supervision with hard labels when training the auxiliary task. 

\noindent \textbf{Domain adaptation or expansion.} Domain adaptation or extension \cite{Li16,Jung18} aims to improve the generalization performance for new domains with existing trained networks. However, unlike in the case of the general domain adaptation problem, SSKT has a significant difference in using the prior knowledge trained in the existing domain for new network training without sharing parameters. In addition, because it does not include any update process for the pretrained network, the update is performed regardless of the performance of the source domain. Moreover, there was no change in the inference performance of the source domain after training.

\section{Self-Supervised Knowledge Transfer}

SSKT supports a training structure that enables knowledge transfer in a variety of scenarios using a CNN. Various scenarios refer to situations not influenced by the network architecture of the target task to be trained and in which the knowledge transfer does not depend on the type of task. We devised a structure that transfers knowledge naturally without compromising the training information of the pretrained network or requiring additional supervision during the target task training process. We achieved this using soft label-based knowledge transfer techniques with auxiliary task learning through self-supervision for the various domains of image recognition variants.

\subsection{SSKT with Single Source}

We define a multi-task network for auxiliary learning, $h_t(x;\theta_t,D_t,T_t)$, where $x$ is the input, $\theta_t$ is a parameter of the target network, $D_t$ is a target dataset, and $T_t$ is the task to be trained. $\theta_t$ is updated simultaneously through the target loss and auxiliary loss during training to solve the primary task. $h_s(x;\theta_s,D_s,T_s)$ describes a source network that receives input $x$ and delivers knowledge to the target network. $\theta_s$ denotes a parameter trained by the source task $T_s$ for the source dataset (ImageNet and Places365) $D_s$. $\theta_s$ is not updated during target task training.
The multi-task network $h_t$ for the primary task training shares up to the top layer of the convolutional block below the primary and auxiliary branches, except for the last feature layer. The last feature layer of the branch for the target task loss outputs $h_{t}^{prim}(x;\theta_t,D_t,T_t)$, and the branch for the auxiliary task loss outputs $h_{t}^{aux}(x;\theta_t,D_t,T_s)$. At this time, the auxiliary task branch could have a transfer module composed of the summation of the bottleneck structure from each convolutional block for effective feature encoding for the auxiliary task. Figure 3 shows an example of a multitask network with a transfer module. The ground truth label of the primary task for the total loss of $h_t$ is given by $y_{t}^{prim}$, for $x$ belonging to $D_t$. The ground truth label for the auxiliary task is given by the softmax output $y_{s}^{aux}$ from $h_s$.

The total loss function for a single-task knowledge transfer based on a multitask network is as follows.

\setlength{\belowdisplayskip}{1pt} \setlength{\belowdisplayshortskip}{1pt}
\setlength{\abovedisplayskip}{1pt} \setlength{\abovedisplayshortskip}{1pt}

\vspace{-3mm}

\begin{align}
  \operatorname*{argmin}_{\theta_t} \Big(L(h_{t}^{prim}(x_i;\theta_t.D_t,T_t),y_{t,i}^{prim}) \notag\\ 
  + \alpha L(h_{t}^{aux}(x_i;\theta_t.D_t,T_s),y_{s,i}^{aux})\Big),
\end{align}

\noindent where $i$ is the $i^{th}$ batch of the training data, $\alpha$ is a balanced parameter for the total loss, and $y_{s,i}^{aux}=h_s(x_i;\theta_s,D_t,T_s )$ is the softmax outputted from the pretrained source network and which conveys the dark knowledge of the pretrained dataset using soft labels. We applied two types of losses of cross-entropy (CE) and knowledge distillation (KD) \cite{Hinton14} for auxiliary task learning:

\vspace{-3mm}

\begin{align}
  L_{CE} = -ylog(c(\frac{g(x;\theta)}{T})), \\
  L_{KD} = KL\Big( c(\frac{g(x;\theta_s)}{T}),c(\frac{g(x;\theta_t)}{T}) \Big),
\end{align}

\noindent where $y$ is a one-hot vector from the ground truth label, $g(x;\theta)$ is the final feature outputted from the network during training, $T$ controls the softening intensity for the softmax output as the temperature parameter, $c$ denotes the softmax function, and $KL$ denotes the KL divergence between the output distributions. The gradient update after $k+1$ iterations is described by

\vspace{-3mm}

\begin{align}
  \theta_{t}^{k+1}=\theta_{t}^{k} - \eta\nabla_{\theta_t}\Big(L(h_{t}^{prim}(x_i;\theta_t.D_t,T_t),y_{t,i}^{prim}) \notag\\ 
  + \alpha L(h_{t}^{aux}(x_i;\theta_t.D_t,T_s),y_{s,i}^{aux}) \Big)
\end{align}

\noindent where $\eta$ is the learning rate. Figure 2(a) shows the structural schematics for the SSKT with a single-source scenario. 
\vspace{1.5mm}

\noindent \textbf{Transfer module (TM).} To encourage the prediction of $y_{s,i}^{aux}$ by $h_t$, we design a bottleneck structure-based transfer module that supports an auxiliary task using the feature outputted from each convolutional block. For an example of a ResNet-based model, each output of the $res$ block is encoded as a fixed-size output by bottleneck and average pooling, and finally, each feature outputted from the transfer bottleneck is summed (see Figure 3).

 \begin{figure}[t!]
\centering
\includegraphics[width=0.8\linewidth]{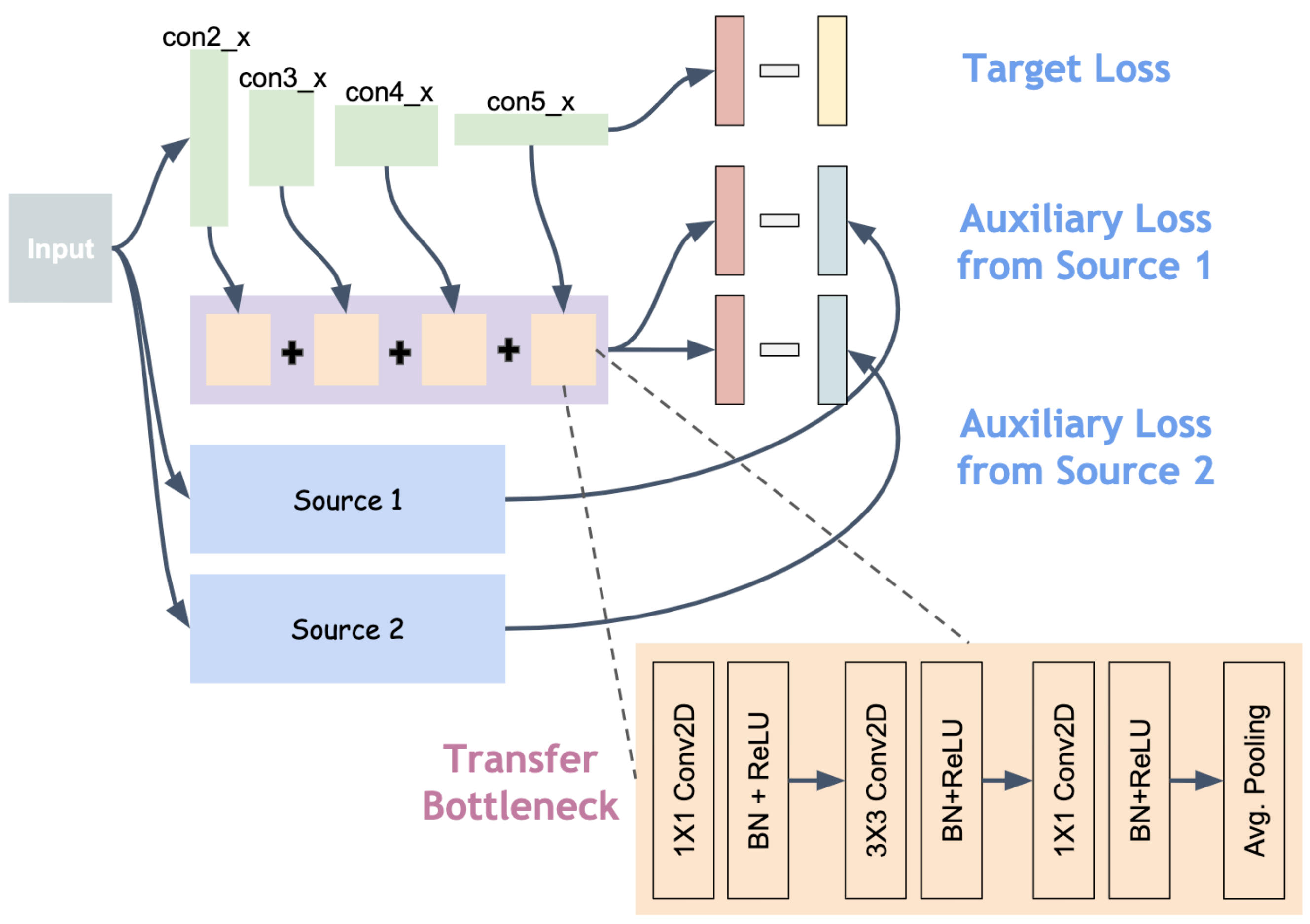}
\vspace{-1mm}
\caption{\textbf{Schematic of transfer modules for efficient auxiliary learning.} The transfer module used in the SSKT consists of the summation of the feature outputs of the bottleneck layers from each convolutional block. The schematic shows an example of the transfer module with ResNet variants for SSKT using multiple sources.}
\label{fig:example}
\vspace{-6mm}
\end{figure}

\subsection{SSKT with Multiple Source}

SSKT can be easily extended to multiple $h_s$ using multiple auxiliary losses. We improved the structure by including $M$ source tasks for knowledge transfer to target tasks, as follows:

\vspace{-4mm}

\begin{align}
  \operatorname*{argmin}_{\theta_t} \Big(L(h_{t}^{prim}(x_i;\theta_t.D_t,T_t),y_{t,i}^{prim}) \notag\\
  + \alpha ( L(h_{t}^{aux}(x_i;\theta_t.D_t,T_{s_1}),y_{s_1,i}^{aux}) \notag\\ 
  + L(h_{t}^{aux}(x_i;\theta_t.D_t,T_{s_2}),y_{s_2,i}^{aux}) + \dots \notag\\
  + L(h_{t}^{aux}(x_i;\theta_t.D_t,T_{s_M}),y_{s_M,i}^{aux}) ) \Big),
\end{align}

\noindent where $s_M$ denotes the index of the source task for knowledge transfer. Theoretically, if the memory is sufficient, we can perform multiple task knowledge transfers from a large number of $M$ source tasks.

\begin{table*}[t!]
\vspace{-3mm}
  \caption{Training details for each task, for knowledge transfer with SSKT. IC denotes image classification, AC denotes action classification, MCIC denotes multi-class image classification, $\beta$ represents momentum, and $\lambda$ represents weight decay. CE denotes cross-entropy, KD denotes knowledge distillation, and BCE denotes binary cross-entropy. In the training model, (s) was trained by learning from scratch, and (f) was trained by fine-tuning.}
  \label{tab:freq}
  \resizebox{\linewidth}{!}{
  \begin{tabular}{c|c|c|c|c|c|c|c}
%    \toprule
\hline
Task ($T$) & Dataset ($D$) & \# cls & Model & Optimizer & LR Scheduler & $\beta$,$\lambda$ & Loss [Task + Auxiliary]\\ \hline
 %   \midrule
IC ($T_t$) & CIFAR10 (C10) & 10 & ResNet[20, 32] & SGD & step (0.1,$[$150:250:350$]$) & 0.9, 5e-4 &[CE + CE or KD]\\ 
		& CIFAR100 (C100) & 100 & ResNet[20, 32] & SGD & step (0.1,$[$60:120:160:200$]$) & 0.9, 5e-4 & [CE + CE or KD]\\ 
				 & STL10 (S10) & 10 & ResNet18[20, 32, 44] & SGD & step (0.1,$[$60:120:160:200$]$) & 0.9, 5e-4 &[CE + CE or KD]\\ 
				 &		&	& MoblieNetV2	& SGD & step (0.1,$[$60:120:160:200$]$) & 0.9, 5e-4 &[CE + CE or KD]\\ 
				 &		&	& DenseNet121	& SGD & step (0.1,$[$60:120:160:200$]$) & 0.9, 5e-4 &[CE + CE or KD]\\ 
    				 & Places365 (P) & 365 & ResNet18 & SGD & step (0.1,$[$30:60:90$]$) & 0.9, 1e-4 &[CE + CE or KD]\\ 
    				 & ImageNet (I) & 1000 & ResNet18 & SGD & step (0.1,$[$30:60:90$]$) & 0.9, 1e-4 &[CE + CE or KD]\\ \hline
MCIC ($T_t$) & Pascal VOC (VOC) & 20 & ResNet[18, 34, 50] (s) & SGD & step (0.1,$[$30:60:90$]$) & 0.9, 1e-4 &[BCE + CE or KD]\\ 
		 	&		& 	 & ResNet18 (f) & SGD & step (0.01,$[$30:60:90$]$) & 0.9, 1e-4 &[BCE + CE or KD]\\\hline
AC ($T_t$) & UCF101 (U101) & 101 & 3D-ResNet18 (s) & SGD & reduce on plateau (0.1) & 0.9, 1e-3 &[CE + CE or KD]\\
	 &  		&  	& 3D-ResNet18 (f) & SGD & reduce on plateau (0.01) & 0.9, 1e-3 &[CE + CE or KD]\\
	 & HMDB51 (H51) & 51 & 3D-ResNet18 (s) & SGD & reduce on plateau (0.1) & 0.9, 1e-3 &[CE + CE or KD]\\
	 & 		& 	 & 3D-ResNet18 (f) & SGD & reduce on plateau (0.01) & 0.9, 1e-3 &[CE + CE or KD]\\ \hline
IC ($T_s$)    		 & Places365 (P) & 365 & ResNet50 & \cite{Zhou18} & \cite{Zhou18} & \cite{Zhou18} & \cite{Zhou18} \\  
    				 & ImageNet (I) & 1000 & ResNet50 & \cite{Paszke19} & \cite{Paszke19} & \cite{Paszke19} & \cite{Paszke19} \\
\hline
\end{tabular}
}
\vspace{-3mm}
\end{table*}

\subsection{SSKT within Different Problem Domains}

Until now, transfers for single and multiple sources have been performed assuming tasks in the same problem domain that do not specify the forms of $D_t$ and $D_s$. We set up a scenario to enable knowledge transfer even when the modalities of $D_t$ and $D_s$ are different or when the problem domains of the tasks are different. In this case, because the characteristics of the input data are different and there is a large difference in the structure of the network for training, the preparation process for knowledge transfer involves data conversion or preprocessing. For knowledge transfer with different problem domains, the total loss function can be defined as:

\vspace{-3mm}

\begin{align}
  \operatorname*{argmin}_{\theta_t} \Big(L(h_{t}^{prim}(x_i;\theta_t.D_t,T_t),y_{t,i}^{prim}) \notag\\ 
  + \alpha L(h_{t}^{aux}(f_s(x_i);\theta_t.D_t,T_s),y_{s,i}^{aux})\Big),
\end{align}

\noindent where the data transformation function $f_s$ converts the data type to match the source task and infer the recognition information to the task of the source domain. For example, if $T_t$ is an action recognition problem using a 3D-CNN, the input $x^{w\times h\times d}\in D_t$ is defined as a 3D tensor. In this case, if a pretrained network for knowledge transfer is obtained through the image recognition problem $T_s$ using 2D-CNN, $f_s:x^{w\times h\times d}\to\hat{x}^{w\times h}$ should be defined as a function that maps a 3D tensor to a 2D matrix into which $h_s$ can be inputted.

Knowledge transfer to other problem domains can also be done with multiple ($M$) source tasks:

\vspace{-3mm}

\begin{align}
  \operatorname*{argmin}_{\theta_t} \Big(L(h_{t}^{prim}(x_i;\theta_t.D_t,T_t),y_{t,i}^{prim}) \notag\\
  + \alpha (  L(h_{t}^{aux}(f_{s_1}(x_i);\theta_t.D_t,T_{s_1}),y_{s_1,i}^{aux}) \notag\\
  + L(h_{t}^{aux}(f_{s_2}(x_i);\theta_t.D_t,T_{s_2}),y_{s_2,i}^{aux}) + \dots \notag\\
  + L(h_{t}^{aux}(f_{s_M}(x_i);\theta_t.D_t,T_{s_M}),y_{s_M,i}^{aux})) \Big),
\end{align}

\noindent in this case, up to $M$ transformation functions can be defined. Gradient updates for knowledge transfer to other problem domains are defined in the same way as single- or multiple-task knowledge transfers between the same domains.

\section{Experimental Results}

To verify the SSKT, we designed a series of experimental scenarios. First, we set a knowledge transfer with a single-source task in the same problem domain. A pretrained CNN on image classification tasks (ImageNet and Places365) was used for knowledge transfer for the same problem domain. Second, we performed knowledge transfer for the image classification problem using multiple source networks. In this case, the target network for target task learning had two or more auxiliary tasks. Third, knowledge transfer scenarios using source networks with different problem domains were tested in two ways. The first method involved target task learning with a different purpose and knowledge transfer to a domain within the same image recognition category. For example, to solve the multiclass classification problem for 2D images, the target network (2D-CNN) has the same type of architecture as the source network (2D-CNN). In this case, knowledge transfer between domains is possible without any special definition of the transformation function $f_s$ for the source network. The second scenario involves knowledge transfer using different problem domains using heterogeneous networks. For example, in training the action recognition network using 3D-CNN, we verified whether the network learned by the image classification problem could transfer the knowledge to the action recognition problem. 

In addition to verifying the knowledge transfer performance of SSKT for each knowledge transfer scenario, the optimization results for the hyperparameters of the loss functions were included. Moreover, our experiments include the relationship between the change in  knowledge transfer performance based on the model architecture and fine tuning, and a performance comparison with other knowledge transfer methods such as KD, DML, and MAXL. All the experimental results included in the table are selected based on the highest performance among combinations of auxiliary loss, transfer module, and source network\footnote{Additional combinations of test results and architecture details can be found in the supplementary materials.}. Table 1 summarizes the experimental settings for the verification of the SSKT.

\subsection{Image Classification $\to$ Image Classification}

\noindent \textbf{SSKT Setting.} To perform SSKT between image classification problems, we used networks pretrained on ImageNet (I) \cite{Deng09} or Places365 (P) \cite{Zhou18} dataset as the source network. The target task datasets were CIFAR10, CIFAR100 \cite{Krizhevsky09}, and STL10 \cite{Coates11}, which contain relatively small amounts of data, as well as large datasets such as ImageNet and Places365. As a pretrained source network, we used the ResNet50 network provided in the PyTorch \textit{torchvision} package \cite{Paszke19}. The transfer module is composed of a bottleneck and average pooling, as shown in Figure 3. Multiple pretrained networks can be utilized in SSKT to apply for multisource task-based knowledge transfer. We used both a ResNet50 model trained on ImageNet and Places365 (P+I) datasets as our source tasks.

\noindent \textbf{Results.} Table 2 shows the performances of learning from scratch and SSKT for the image classification problem on CIFAR10 (C10), 100 (C100), and STL10 (S10) datasets. For knowledge transfer between image classification problems, the use of SSKT improves the generalization performance on all the tested datasets. The performance change when the transfer module is applied to efficiently transform the features for auxiliary task learning is also evident. Notably, the highest improvement in the generalization performance was on the SLT10 dataset, which contains a small proportion of the training data. This means that knowledge transfer using SSKT is more effective when the amount of supervision is relatively low. Table 3 shows the performance changes of the SSKT on large-scale image datasets. When SSKT was applied to both Place365 and ImageNet, there was a steady performance improvement. In particular, the highest performance improvement was obtained when the source networks were used simultaneously (P + I). However, for large datasets, it was difficult to confirm the effect of improving the performance of the transfer module.

\begin{table}[t!]
\vspace{-3mm}
  \caption{Performance change of SSKT for a single source compared to training from scratch. The best-performing model for each dataset is highlighted in bold. All the experiments evaluated the test performance thrice from the same random seed for the model. TM denotes transfer module, and R[depth] denotes ResNet structure.}
  \label{tab:freq}
  \resizebox{\linewidth}{!}{
  \begin{tabular}{c|c|c|c|c|c|c}
%    \toprule
\hline
   $T_s$ 	& 	$T_t$ 	& Model & Method & TM & Loss & acc.\\ \hline
 %   \midrule 
 	- 	& C10 		& R20 & scratch & - & CE & 92.19$\pm$0.09\\ 
P 		&  			& R20 & SSKT & o & CE+KD & 92.25$\pm$0.04\\ 
I 		&  			& R20 & SSKT & o & CE+CE & 92.44$\pm$0.05\\ 
P+I		& 			& R20 & SSKT & x & CE+KD & \textbf{92.46$\pm$0.15}\\ \hline
 	- 	& 	 		& R32 & scratch & - & CE & 93.21$\pm$0.09\\ 
P 		&  			& R32 & SSKT & x & CE+KD & 92.87$\pm$0.31\\ 
I 		&  			& R32 & SSKT & x & CE+CE & 93.26$\pm$0.08\\ 
P+I		& 			& R32 & SSKT & o & CE+CE & \textbf{93.38$\pm$0.02}\\ \hline \hline
 	- 	& C100 		& R20 & scratch & - & CE & 68.26$\pm$0.36\\ 
P 		&  			& R20 & SSKT & x & CE+KD & 68.01$\pm$0.42\\ 
I 		&  			& R20 & SSKT & o & CE+CE & \textbf{68.63$\pm$0.12}\\ 
P+I		& 			& R20 & SSKT & o & CE+CE & 68.56$\pm$0.23\\ \hline
 	- 	& 	 		& R32 & scratch & - & CE & 70.33$\pm$0.19\\ 
P 		&  			& R32 & SSKT & x & CE+CE & 69.97$\pm$0.16\\ 
I 		&  			& R32 & SSKT & o & CE+CE & 70.75$\pm$0.06\\ 
P+I		& 			& R32 & SSKT & o & CE+CE & \textbf{70.94$\pm$0.36}\\ \hline \hline
 	- 	& S10 		& R20 & scratch & - & CE & 81.15$\pm$0.34\\ 
P	 	&  			& R20 & SSKT & o & CE+CE & 82.76$\pm$0.05\\
I   		&			& R20 & SSKT & o & CE+CE & 83.45$\pm$0.07\\ 
P+I		&  			& R20 & SSKT & o & CE+CE & \textbf{84.56$\pm$0.35}\\ \hline
 	- 	& 	 		& R32 & scratch & - & CE & 81.19$\pm$0.17\\ 
P	 	&  			& R32 & SSKT & o & CE+CE & 83.06$\pm$0.27\\ 
I		&			& R32 & SSKT & o & CE+CE & \textbf{83.68$\pm$0.28}\\ 
P+I		&  			& R32 & SSKT & o & CE+CE & 83.4$\pm$0.2\\ \hline
 	- 	& 	 		& R44 & scratch & - & CE & 80.18$\pm$0.54\\ 
P	 	&  			& R44 & SSKT & o & CE+CE & 82.68$\pm$0.39\\ 
I		&			& R44 & SSKT & o & CE+CE & \textbf{83.59$\pm$0.13}\\ 
P+I		&  			& R44 & SSKT & o & CE+CE & 83.44$\pm$0.15\\ \hline
\end{tabular}
}
\vspace{-5mm}
\end{table}

\begin{table}[t!]
  \caption{Performance change when applying the SSKT and training from scratch on a large-scale image dataset.}
  \label{tab:freq}
  \resizebox{\linewidth}{!}{
  \begin{tabular}{c|c|c|c|c|c|c}
%    \toprule
\hline
   $T_s$ 	& 	$T_t$ 	& Model & Method & TM & Loss & acc.\\ \hline
 %   \midrule 
 	- 	& P	 		& R18 & scratch & - & CE & 50.92\\ 
P 		&  			& R18 & SSKT & o & CE+CE & 54.5\\ 
I 		&  			& R18 & SSKT & o & CE+CE & 53.67\\ 
P+I		& 			& R18 & SSKT & x & CE+CE & \textbf{54.78}\\ \hline
 	- 	& I 			& R18 & scratch & - & CE & 64.14\\ 
P 		&  			& R18 & SSKT & o & CE+CE & 64.99\\ 
I 		&  			& R18 & SSKT & x & CE+CE & 67.79\\ 
P+I		& 			& R18 & SSKT & x & CE+CE & \textbf{70.57}\\ \hline
\end{tabular}
}
\vspace{-3mm}
\end{table}

\vspace{-2mm}

\subsection{Image Classification $\to$ Multi-class Image Classification}

\noindent \textbf{SSKT Setting.} To verify whether knowledge transfer using SSKT is effective between different problem domains, knowledge transfer between image classification and multiclass image classification was performed. We tested multiclass image classification using the PASCAL VOC dataset \cite{Everingham15} as the target task. The target loss function for solving the multiclass image classification was used as the binary cross-entropy (BCE). Table 1 lists the training details for the target and source tasks.

\noindent \textbf{Results.} Table 4 shows the performance of learning from scratch for multiclass image classification and the performance change when SSKT is included in the training process. For the PASCAL VOC multi-class classification task, we used the standard average precision (AP) as the accuracy measurement to evaluate the predictions. The use of SSKT significantly improved the knowledge transfer performance between different problem domains. The improvement was greater than that for knowledge transfer between the same problem domains. In addition, we verified an additional performance improvement when applying SSKT to multi-source networks.

\begin{table}[t!]
\vspace{-1mm}
  \caption{Results of SSKT from the image classification problem to the multi-class image classification problem.}
  \label{tab:freq}
  \resizebox{\linewidth}{!}{
  \begin{tabular}{c|c|c|c|c|c|c}
%    \toprule
\hline
   $T_s$ 	& 	$T_t$ 	& Model & Method & TM & Loss & acc. \\ \hline
 %   \midrule 
 	- 	& VOC	 	& R18 & scratch & - & BCE & 67.28$\pm$0.25\\ 
P 		&  			& R18 & SSKT & o & BCE+CE & 74.76$\pm$0.17\\ 
I 		&  			& R18 & SSKT & x & BCE+CE & 74.78$\pm$0.09\\ 
P+I		& 			& R18 & SSKT & o & BCE+CE & \textbf{76.42$\pm$0.06}\\ \hline
 	- 	& 		 	& R34 & scratch & - & BCE & 66.0$\pm$0.49\\ 
P 		&  			& R34 & SSKT & o & BCE+CE & 75.65$\pm$0.12\\ 
I 		&  			& R34 & SSKT & o & BCE+CE & 75.14$\pm$0.14\\ 
P+I		& 			& R34 & SSKT & o & BCE+CE & \textbf{77.02$\pm$0.02}\\ \hline
 	- 	& 		 	& R50 & scratch & - & BCE & 61.16$\pm$0.34\\ 
P 		&  			& R50 & SSKT & o & BCE+CE & 74.44$\pm$0.06\\ 
I 		&  			& R50 & SSKT & o & BCE+CE & 74.24$\pm$0.05\\ 
P+I		& 			& R50 & SSKT & o & BCE+CE & \textbf{77.1$\pm$0.14}\\ \hline
\end{tabular}
}
\vspace{-5mm}
\end{table}

\subsection{Image Classification $\to$ Action Classification}

\noindent \textbf{SSKT Setting.} Although there is a difference between problem domains, the performance improvement of knowledge transfer may be relatively predictable because the modalities of the input data of the source and target tasks are the same. We used a pretrained image classification network as the source task and set the target task as an action recognition problem. We employed 3D-CNN to design an experiment for a more extended knowledge transfer. To utilize the dark knowledge of the source network as described in Equation (6), we need to define a transformation function $f_s:x^{w\times h\times d}\to\hat{x}^{w\times h}$. Because SSKT is not designed to define good $f_s$, we simply limited the role of $f_s$ in extracting the center frame from a 3D video clip. As a transfer module for inferring auxiliary tasks in a 3D-CNN, the output features of the target model were further processed using a 3D bottleneck structure. We used a 3D ResNet-based baseline model \cite{Hara18} to train the action classification networks. Table 1 lists the training details for 3D ResNet and source networks.

\noindent \textbf{Results.} Table 5 shows the performance of learning from scratch with the 3D-ResNet on the UCF101 \cite{Soomro12} and HMDB51 \cite{Kuehne11} datasets, as well as the performance change when SSKT was used in the training process. All the results were obtained from \textit{split 1} for each dataset. Note that the auxiliary task in SSKT prevents overfitting during the training of the 3D-CNN model. Similar to other SSKT settings, the performance improvement was highest when using a multi-source network. We further note that the use of SSKT between heterogeneous problem domains can improve the performance further, depending on the definition of $f_s$.

% \vspace{-2mm}
\subsection{Further Analysis and Discussion}

\noindent \textbf{Parameter optimization.} We performed a series of experiments for the temperature parameter $T$ included in the auxiliary loss, the balance parameter $\alpha$ included in the total loss, and the presence of the transfer module. Figure 4 shows the graphs of the parameter optimization results on the STL10 and PASCAL VOC datasets. For STL10, the use of a transfer module in the SSKT process yielded a high performance in most cases; however, in the case of PASCAL VOC, the transfer module did not play a significant role.

 \begin{figure*}[t!]
\centering
\includegraphics[width=1.0\linewidth]{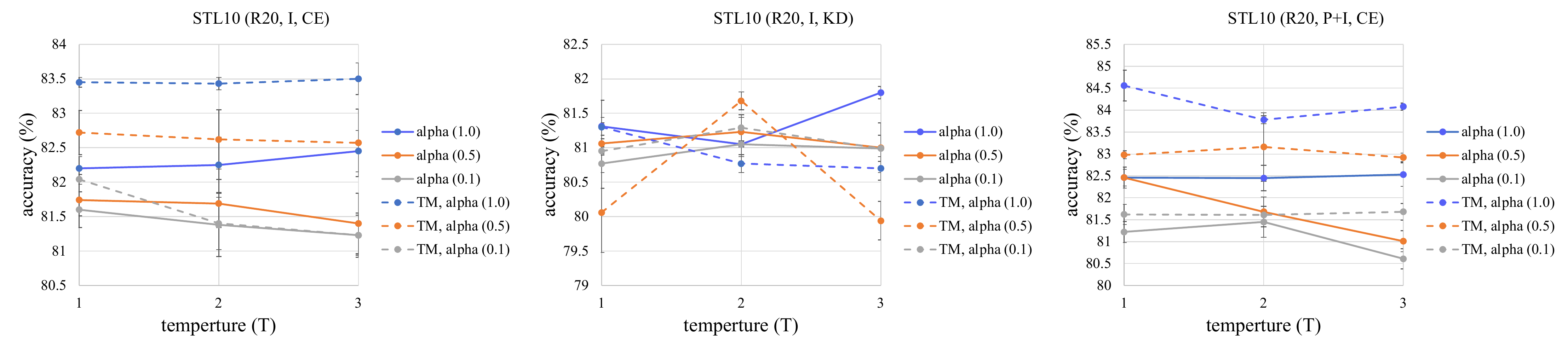}
\vspace{-5mm}
\caption{\textbf{Hyper parameter optimization of SSKT.} The title of each graph is composed of $D_t$ (target model, $T_s$, auxiliary loss). $T$ is the temparture parameter of each auxiliary loss, and $\alpha$ is the balance parameter of the total loss.}
\label{fig:example}
\vspace{-3mm}
\end{figure*}

\noindent \textbf{Model comparison.} We applied SSKT to MobileNet V2 (MV2) \cite{Sandler18} and DenseNet121 (D121) \cite{Huang17} to evaluate the model performance in CNN architectures other than ResNet variants. The source network used the ResNet50 model, which was previously trained on Places365 and ImageNet, similar to other SSKT settings. In evaluating the STL10 dataset, Table 6 shows that MobileNet V2 and DenseNet121 exhibit performance improvements with the SSKT. The results in the table show the highest performance in the configuration among single-and multi-source SSKT settings.

\begin{table}[t!]
\vspace{-3mm}
  \caption{Results of SSKT from image classification to action classification with different modalities of the input data.}
  \label{tab:freq}
  \resizebox{\linewidth}{!}{
  \begin{tabular}{c|c|c|c|c|c|c}
%    \toprule
\hline
   $T_s$ 	& 	$T_t$ 	& Model & Method & TM & Loss & acc.\\ \hline
 %   \midrule 
 	- 	& U101	 	& 3DR18 & scratch & - & CE & 43.28\\ 
P 		&  			& 3DR18 & SSKT & o & CE+CE & 45.35\\ 
I 		&  			& 3DR18 & SSKT & x & CE+CE & 46.62\\ 
P+I		& 			& 3DR18 & SSKT & x & CE+CE & \textbf{52.19}\\ \hline \hline
 	- 	& H51	 	& 3DR18 & scratch & - & CE & 17.14\\ 
P 		&  			& 3DR18 & SSKT & o & CE+CE & 18.77\\ 
I 		&  			& 3DR18 & SSKT & o & CE+KD & 18.77\\ 
P+I		& 			& 3DR18 & SSKT & o & CE+CE & \textbf{20.54}\\ \hline
\end{tabular}
}
\vspace{-3mm}
\end{table}

\begin{table}[t!]
  \caption{SSKT results using MobileNet V2 (MV2) and Dense-Net121 (D121).}
  \label{tab:freq}
  \resizebox{\linewidth}{!}{
  \begin{tabular}{c|c|c|c|c|c|c}
%    \toprule
\hline
   $T_s$ 	& 	$T_t$ 	& Model & Method & TM & Loss & acc.\\ \hline
 %   \midrule 
 	- 	& S10	 	& R20 & scratch & - & CE & 81.15$\pm$0.34\\ 
 	- 	& 		 	& MV2 & scratch & - & CE & 72.26$\pm$0.83\\ 
 	- 	& 		 	& D121 & scratch & - & CE & 72.02$\pm$0.48\\ 	
P+I 		&  			& R20 & SSKT & o & CE+ CE& \textbf{84.56$\pm$0.35}\\ 
P+I		&  			& MV2 & SSKT & o & CE+CE & \textbf{76.96$\pm$0.39}\\ 
P+I		& 			& D121 & SSKT & x & CE+CE & \textbf{77.03$\pm$0.17}\\ \hline 
\end{tabular}
}
\vspace{-3mm}
\end{table}

\noindent \textbf{Relation to fine-tuning.} In SSKT scenarios, fine-tuning and learning from scratch can be applied for target network training. We adopted fine-tuning-based SSKT with a ResNet18 model trained in ImageNet for a multiclass image classification problem using the PASCAL VOC dataset. In addition, using the pretrained 3D-ResNet18 model on the Kinetics-400 dataset \cite{Kay17}, SSKT was applied to the action classification problem for the UCF101 and HMDB51 datasets. Table 7 shows that even if fine-tuning based on predetermined weights is performed, the performance is improved in all the experimental settings. However, unlike the evaluation results in the case of training from scratch, the performance improvement from multiple sources was not significant, and the best performance was obtained when the KD loss was used.

\begin{table}[t!]
  \caption{SSKT results using pretrained weights. \textit{ft} denotes fine-tuning and K denotes Kinetics-400 dataset.}
  \label{tab:freq}
  \resizebox{\linewidth}{!}{
  \begin{tabular}{c|c|c|c|c|c|c}
%    \toprule
\hline
   $T_s$ 	& 	$T_t$ 	& Model & Method & TM & Loss & acc.\\ \hline
 %   \midrule 
 	- 	& VOC	 	& R18 & \textit{ft} (I) & - & CE & 90.52$\pm$0.11\\ 
P 		&  			& R18 & SSKT & x & CE+KD & \textbf{92.28$\pm$0.06}\\ 
I 		&  			& R18 & SSKT & x & CE+KD & 92.26$\pm$0.07\\ 
P+I		& 			& R18 & SSKT & o & CE+KD & 92.25$\pm$0.07\\ \hline \hline
 	- 	& U101	 	& 3DR18 & \textit{ft} (K) & - & CE & 83.95\\ 
P 		&  			& 3DR18 & SSKT & x & CE+KD & \textbf{84.58}\\ 
I 		&  			& 3DR18 & SSKT & o & CE+KD & 84.37\\ 
P+I		& 			& 3DR18 & SSKT & o & CE+KD & 84.19\\ \hline \hline
 	- 	& H51	 	& 3DR18 & \textit{ft} (K) & - & CE & 56.64\\ 
P 		&  			& 3DR18 & SSKT & o & CE+KD & \textbf{57.82}\\ 
I 		&  			& 3DR18 & SSKT & o & CE+KD & 57.75\\ 
P+I		& 			& 3DR18 & SSKT & o & CE+CE & 57.29\\ \hline
\end{tabular}
}
\vspace{-3mm}
\end{table}

\noindent \textbf{Comparison with other knowledge transfer methods.} For a further analysis of SSKT, we compared its performance with those of typical knowledge transfer methods, namely KD and DML. For KD, the details for learning were set the same as in \cite{Hinton14}, and for DML, training was performed in the same way as in \cite{Zhang18}. Table 8 shows the evaluation performance in terms of knowledge transfer. In the case of 3D-CNN-based action classification, KD was not performed, and the network used for DML had the model with the same conditions as the source and target networks of SSKT. In the case of DML for 3D-CNN, training was performed by replacing ResNet50 with 3D-ResNet50 under the same conditions. In the case of action classification, both learning from scratch and fine-tuning results were included. In most cases, under similar training conditions, SSKT exhibited a higher generalization performance than other knowledge transfer techniques. Moreover, training in problem domains other than the image classification problem for DML was difficult. In particular, in the case of multiclass classification, neither network was sufficiently trained. SSKT achieved improved performance in all problem domains with multiple experimental setups. 

Table 9 shows the performance comparison with MAXL, another auxiliary learning-based transfer learning method \cite{Liu19}. In addition to the target task's class label, MAXL requires a manually defined hierarchical auxiliary class structure ($\psi[i]=2,3,5,10$). We compared the performance on the CIFAR10 dataset by applying the same four-stage hierarchical auxiliary class label as defined in \cite{Liu19}. Despite not requiring additional class labels, SSKT showed better recognition performance than MAXL in all settings. In particular, the same learning scheduler in \cite{Liu19} showed better performance than MAXL (comparisons with VGG16 in Table 9), and SSKT achieved a larger performance gap when changing the source task and hyper-parameter setting.

\begin{table}[t!]
% \vspace{-3mm}
  \caption{Comparison with other knowledge transfer methods. (s) denote learning from scratch and (f) denote fine-tuning.}
  \label{tab:freq}
  \resizebox{\linewidth}{!}{
  \begin{tabular}{c|c|c|c|c}
%    \toprule
\hline
$T_t$  & Model & KD & DML & SSKT ($T_s$) \\ \hline
 %   \midrule 
C10 		& R20 & 91.75$\pm$0.24 & 92.37$\pm$0.15 & \textbf{92.46$\pm$0.15} (P+I) \\ 
		& R32 & 92.61$\pm$0.31 & 93.26$\pm$0.21 & \textbf{93.38$\pm$0.02} (P+I) \\ \hline
C100	& R20 & 68.66$\pm$0.24 & \textbf{69.48$\pm$0.05} & 68.63$\pm$0.12 (I) \\ 
		& R32 & 70.5$\pm$0.05 	& \textbf{71.9$\pm$0.03} & 70.94$\pm$0.36 (P+I) \\ \hline
S10 		& R20 & 77.67$\pm$1.41 & 78.23$\pm$1.23 & \textbf{84.56$\pm$0.35} (P+I) \\ 
		& R32 & 76.07$\pm$0.67 & 77.14$\pm$1.64 & \textbf{83.68$\pm$0.28} (I) \\ \hline
VOC		& R18 & 64.11$\pm$0.18 & 39.89$\pm$0.07 & \textbf{76.42$\pm$0.06} (P+I) \\ 
		& R34 & 64.57$\pm$0.12 & 39.97$\pm$0.16 & \textbf{77.02$\pm$0.02} (P+I) \\ 
		& R50 & 62.39$\pm$0.6 & 39.65$\pm$0.03 & \textbf{77.1$\pm$0.14} (P+I) \\ \hline
U101	& 3DR18 (s) & - & 13.8 & \textbf{52.19} (P+I)\\ 
		& 3DR18 (f) & - & 83.95 & \textbf{84.58} (P) \\ \hline
H51		& 3DR18 (s) & - & 3.01 & \textbf{17.91} (P+I) \\ 
		& 3DR18 (f) & - & 56.44 & \textbf{57.82} (P) \\ \hline
\end{tabular}
}
\vspace{-4mm}
\end{table}

\begin{table}[t!]
  \caption{Comparison with MAXL. V16 denotes VGG16 \cite{Simonyan15}. F denotes focal loss 
  \cite{Liu19}}
  \label{tab:freq}
  \resizebox{\linewidth}{!}{
  \begin{tabular}{c|c|c|c|c}
%    \toprule
\hline
$T_t$  & Model (Loss) & MAXL ($\psi[i]$) & SSKT ($T_s$) & $T_s$ Model  \\ \hline
 %   \midrule 
C10 		& V16 (F) & 93.27$\pm$0.09 (2) &  94.1$\pm$0.1 (I) & V16 \\ 
		& V16 (F) & 93.47$\pm$0.08 (3) &  92.72$\pm$0.15 (P) & R50 \\ 
		& V16 (F) & 93.49$\pm$0.05 (5) &  93.06$\pm$0.2  (P+I) & R50, V16 \\ 
		& V16 (F) & 93.10$\pm$0.08 (10) &  92.54$\pm$0.17(P+I) & R50, R50 \\ 
		& V16 (CE) & - &  \textbf{94.22$\pm$0.02} (I) & V16 \\ 
		& V16 (CE) & - &  93.12$\pm$0.12 (P) & R50 \\  
		& V16 (CE) & - &  93.67$\pm$0.17 (P+I) & R50, V16 \\
		& V16 (CE) & - &  93.26$\pm$0.12 (P+I) & R50, R50 \\ \hline
		& R20 (F) & 91.53$\pm$0.33 (2)&  91.48$\pm$0.03 (I) & V16 \\ 
		& R20 (F) & 91.52$\pm$0.1 (3)&  91.26$\pm$0.17 (P) & R50 \\ 
		& R20 (F) & 91.38$\pm$0.47 (5)&  90.93$\pm$0.01 (P+I) & R50, V16 \\ 
		& R20 (F) & 91.56$\pm$0.16 (10)&  91.11$\pm$0.18 (P+I) & R50, R50 \\ 
		& R20 (CE) & - &  92.44$\pm$0.05 (I) & R50 \\ 
		& R20 (CE) & - &  92.25$\pm$0.04 (P) & R50\\ 
		& R20 (CE) &  -&  \textbf{92.46$\pm$0.15} (P+I) & R50, R50 \\ \hline		
\end{tabular}
}
% \vspace{-6mm}
\end{table}

\vspace{-2mm}
\section{Conclusions}

We proposed a simple yet powerful knowledge transfer method, namely SSKT, that enables knowledge transfer between heterogeneous networks and datasets. We validated the SSKT using several knowledge transfer scenarios based on a CNN. For further investigations of the SSKT, depending on the target task where knowledge transfer takes place, additional considerations for the transfer module design must be made. Moreover, the following experiments are required to verify how the performance of the target model varies with the architecture of the source model trained on the same dataset. Variations in the data transformation function $f_s$ for the source networks may be another avenue worth pursuing. Further, there is a need to perform analyses, such as the one proposed in \cite{Du19}, to analyze the factors involved when the backpropagation of the auxiliary task loss affects the performance of the target task. We expect that further research will follow, as SSKT may provide a different perspective on knowledge transfer. \\

\noindent \textbf{Acknowledgement.}  This work was supported by the Korea Medical Device Development Fund grant funded by the Korea government (the Ministry of Science and ICT, the Ministry of Trade, Industry and Energy, the Ministry of Health Welfare, the Ministry of Food and Drug Safety) (Project Number: 202012A02-02).

{\small
\bibliographystyle{ieee_fullname}
\bibliography{egbib}

\begin{thebibliography}{10}\itemsep=-1pt

\bibitem{Chen20_2}
T. Chen, S. Kornblith, M. Norouzi, and G. Hinton.
\newblock Big self-supervised models are strong semi-supervised learners.
\newblock In {\em In Proc. of NeurIPS}, 2020.

\bibitem{Chen20_1}
T. Chen, S. Kornblith, M. Norouzi, and G. Hinton.
\newblock A simple framework for contrastive learning of visual
  representations.
\newblock In {\em In Proc. of ICML}, 2020.

\bibitem{Chung20}
I. Chung, S. Park, J. Kim, and N. Kwak.
\newblock Feature map-level online adversarial knowledge distillation.
\newblock In {\em In Proc. of ICML}, 2020.

\bibitem{Coates11}
A. Coates, A. Ng, and H. Lee.
\newblock An analysis of single-layer networks in unsupervised feature
  learning.
\newblock In {\em In Proc. of AISTAT}, 2011.

\bibitem{Deng09}
J. Deng, W. Dong, R. Socher, L.~J. Li, K. Li, and L. Fei-Fei.
\newblock Imagenet: A large-scale hierarchical image database.
\newblock In {\em In Proc. of CVPR}, 2009.

\bibitem{Du19}
Y. Du, W.~M. Czarnecki, S.~M. Jayakumar, R. Pascanu, and B. Lakshminarayanan.
\newblock Adapting auxiliary losses using gradient similarity.
\newblock In {\em In Proc. of NeurIPSW}, 2019.

\bibitem{Everingham15}
M. Everingham, S.~M.~A. Eslami, L.~V. Gool, C.~K.~I. Williams, Winn J, and A.
  Zisserman.
\newblock The pascal visual object classes challenge - a retrospective.
\newblock {\em International Journal of Computer Vision}, 111(1):98--136, 2015.

\bibitem{Furlanello18}
T. Furlanello, Z.~C. Lipton, M. Tschannen, L. Itti, and A. Anandkumar.
\newblock Born again neural networks.
\newblock In {\em In Proc. of ICML}, 2018.

\bibitem{Goyal19}
P. Goyal, D. Mahajan, A. Gupta, and I. Misra.
\newblock A simple framework for contrastive learning of visual
  representations.
\newblock In {\em In Proc. of ICCV}, 2019.

\bibitem{Hara18}
K. Hara, H. Kataoka, and Y. Satoh.
\newblock Can spatiotemporal 3d cnns retrace the history of 2d cnns and
  imagenet?
\newblock In {\em In Proc. of CVPR}, 2018.

\bibitem{He20}
K. He, H. Fan, Y. Wu, S. Xie, and R. Girshick.
\newblock Momentum contrast for unsupervised visual representation learning.
\newblock In {\em In Proc. of CVPR}, 2020.

\bibitem{He19}
K. He, R. Girshick, and P. Dollár.
\newblock Rethinking imagenet pre-training.
\newblock In {\em In Proc. of ICCV}, 2019.

\bibitem{He16}
K. He, X. Zhang, S. Ren, and J. Sun.
\newblock Deep residual learning for image recognition.
\newblock In {\em In Proc. of CVPR}, 2016.

\bibitem{Hinton14}
G. Hinton, O. Vinyals, and J. Dean.
\newblock Distilling the knowledge in a neural network.
\newblock In {\em In Proc. of NIPSW}, 2014.

\bibitem{Huang17}
G. Huang, Z. Liu, L. van~der Maaten, and K.~Q. Weinberger.
\newblock Densely connected convolutional networks.
\newblock In {\em In Proc. of CVPR}, 2017.

\bibitem{Jung18}
H. Jung, J. Ju, M. Jung, and J. Kim.
\newblock Less-forgetful learning for domain expansion in deep neural networks.
\newblock In {\em In Proc. of AAAI}, 2018.

\bibitem{Kay17}
W. Kay, J. Carreira, K. Simonyan, B. Zhang, C. Hillier, S. Vijayanarasimhan, F.
  Viola, T. Green, T. Back, P. Natsev, M. Suleyman, and A. Zisserman.
\newblock The kinetics human action video dataset.
\newblock In {\em \textit{CoRR} abs/1705.06950}, 2017.

\bibitem{Kim20}
J. Kim, M. Hyun, I. Chung, and N. Kwak.
\newblock Feature fusion for online mutual knowledge distillation.
\newblock In {\em In Proc. of ICPR}, 2020.

\bibitem{Krizhevsky09}
A. Krizhevsky.
\newblock Learning multiple layers of features from tiny images.
\newblock In {\em Technical Report}, 2009.

\bibitem{Krizhevsky12}
A. Krizhevsky, I. Sutskever, and G.~E. Hinton.
\newblock Imagenet classification with deep convolutional neural networks.
\newblock In {\em In Proc. of NIPS}, 2012.

\bibitem{Kuehne11}
H. Kuehne, H. Jhuang, E. Garrote, T.~A. Poggio, and T. Serre.
\newblock Hmdb: a large video database for human motion recognition.
\newblock In {\em In Proc. of ICCV}, 2011.

\bibitem{Li16}
Z. Li and D. Hoiem.
\newblock Learning without forgetting.
\newblock In {\em In Proc. of ECCV}, 2016.

\bibitem{Liu19}
S. Liu, A.~J. Davison, and E. Johns.
\newblock Self-supervised generalisation with meta auxiliary learning.
\newblock In {\em In Proc. of NeurIPS}, 2019.

\bibitem{Liu16}
W. Liu, D. Anguelov, D. Erhan, C. Szegedy, S. Reed, C.~Y. Fu, and A.~C. Berg.
\newblock Ssd: Single shot multibox detector.
\newblock In {\em In Proc. of ECCV}, 2016.

\bibitem{Long15}
J. Long, E. Shelhamer, and T. Darrell.
\newblock Fully convolutional networks for semantic segmentation.
\newblock In {\em In Proc. of CVPR}, 2015.

\bibitem{Muller20}
R. Müller, S. Kornblith, and G. Hinton.
\newblock Subclass distillation.
\newblock In {\em \textit{CoRR} abs/2002.03936}, 2020.

\bibitem{Park19}
W. Park, D. Kim, Y. Lu, and M. Cho.
\newblock Imagenet classification with deep convolutional neural networks.
\newblock In {\em In Proc. of CVPR}, 2019.

\bibitem{Paszke19}
A. Paszke, S. Gross, F. Massa, A. Lerer, J. Bradbury, G. Chanan, T. Killeen, Z.
  Lin, N. Gimelshein, L. Antiga, A. Desmaison, A. Köpf, E. Yang, Z. DeVito, M.
  Raison, A. Tejani, S. Chilamkurthy, B. Steiner, L. Fang, J. Bai, and S.
  Chintala.
\newblock Pytorch: An imperative style, high-performance deep learning library.
\newblock In {\em In Proc. of NeurIPS}, 2019.

\bibitem{Ren15}
S. Ren, K. He, R. Girshick, and J. Sun.
\newblock Faster r-cnn: Towards real-time object detection with region proposal
  networks.
\newblock In {\em In Proc. of NIPS}, 2015.

\bibitem{Romero15}
A. Romero, N. Ballas, S.~E. Kahou, A. Chassang, C. Gatta, and Y. Bengio.
\newblock Fitnets: Hints for thin deep nets.
\newblock In {\em In Proc. of ICLR}, 2015.

\bibitem{Sandler18}
M. Sandler, A. Howard, M. Zhu, A. Zhmoginov, and L.~C. Chen.
\newblock Mobilenetv2: Inverted residuals and linear bottlenecks.
\newblock In {\em In Proc. of CVPR}, 2018.

\bibitem{Ramprasaat17}
R.~R. Selvaraju, M. Cogswell, A. Das, R. Vedantam, D. Parikh, and D. Batra.
\newblock Grad-cam: Visual explanations from deep networks via gradient-based
  localization.
\newblock In {\em In Proc. of ICCV}, 2017.

\bibitem{Simonyan15}
Karen Simonyan and Andrew Zisserman.
\newblock Very deep convolutional networks for large-scale image recognition.
\newblock In {\em In Proc. of ICLR}, 2015.

\bibitem{Soomro12}
K. Soomro, A.~R. Zamir, and M. Shah.
\newblock Ucf101: A dataset of 101 human actions classes from videos in the
  wild.
\newblock In {\em \textit{CoRR} abs/1212.0402}, 2012.

\bibitem{Tan18}
C. Tan, F. Sun, T. Kong, W. Zhang, C. Yang, and C. Liu.
\newblock A survey on deep transfer learning.
\newblock In {\em In Proc. of ICANN}, 2018.

\bibitem{Wu20}
G. Wu and S. Gong.
\newblock Peer collaborative learning for online knowledge distillation.
\newblock In {\em In Proc. of CVPR}, 2020.

\bibitem{Yosinski14}
J. Yosinski, Clune J, Bengio Y, and Lipson H.
\newblock How transferable are features in deep neural networks?
\newblock In {\em In Proc. of NIPS}, 2014.

\bibitem{Zamir18}
A. Zamir, A. Sax, W. Shen, L. Guibas, J. Malik, and S. Savarese.
\newblock Taskonomy: Disentangling task transfer learning.
\newblock In {\em In Proc. of CVPR}, 2018.

\bibitem{Zamir20}
A. Zamir, A. Sax, T. Yeo, O. Kar, N. Cheerla, R. Suri, Z. Cao, J. Malik, and L.
  Guibas.
\newblock Robust learning through cross-task consistency.
\newblock In {\em In Proc. of CVPR}, 2020.

\bibitem{Zhang18}
Y. Zhang, T. Xiang, T.~M. Hospedales, and H. Lu.
\newblock Deep mutual learning.
\newblock In {\em In Proc. of CVPR}, 2018.

\bibitem{Zhou18}
B. Zhou, A. Lapedriza, A. Khosla, A. Oliva, and A. Torralba.
\newblock Places: A 10 million image database for scene recognition.
\newblock {\em IEEE Transactions on Pattern Analysis and Machine Intelligence},
  40(6):1452--1464, 2018.

\end{thebibliography}
}

\section*{Supplementary Materials}

\section*{A. SSKT with Multiple Tasks and Different Problem Domains}
Self-supervised knowledge transfer (SSKT) supports a training procedure that enables effective transfer learning in a variety of scenarios using deep neural networks. SSKT has a structure that transfers pretrained knoweldge naturally, without compromising the training information of the pretrained network or requiring additional supervision in the target task training process. We achieved this goal using the soft label-based knowledge transfer techniques with auxiliary learning through self-supervision, for the various domain of image recognition variants. Final formulation of the SSKT as follows:

\begin{align}
  \operatorname*{argmin}_{\theta_t} \Big(L(h_{t}^{prim}(x_i;\theta_t.D_t,T_t),y_{t,i}^{prim}) \notag\\
  + \alpha (  L(h_{t}^{aux}(f_{s_1}(x_i);\theta_t.D_t,T_{s_1}),y_{s_1,i}^{aux}) \notag\\
  + L(h_{t}^{aux}(f_{s_2}(x_i);\theta_t.D_t,T_{s_2}),y_{s_2,i}^{aux}) + \dots \notag\\
  + L(h_{t}^{aux}(f_{s_M}(x_i);\theta_t.D_t,T_{s_M}),y_{s_M,i}^{aux})) \Big).
\end{align}

\noindent We define a multi-task network $h_t (x;\theta_t,D_t,T_t)$, where $x$ is the input, $\theta_t$ is a parameter of the target network, $D_t$  is a target dataset, and $T_t$  is the task to be trained. $\theta_t$  is updated simultaneously through target loss and auxiliary loss during training to solve the primary task. $h_s (x;\theta_s,D_s,T_s)$ denotes a source network that receives the input $x$ and delivers knowledge to the target network. $\theta_s$ denotes a parameter trained by the source task $T_s$ for the source data set $D_s$. $\theta_s$ is not updated during the target task training. $i$ is the $i^{th}$ batch of the training data, $\alpha$ is balanced parameter for total loss, and $y_{s,i}^{aux}=h_s(x_i;\theta_s,D_s,T_s )$ is the softmax output from the pretrained source network and conveys the dark knowledge of the pretrained dataset by soft labels. The data transformation function $f_s$ converts the data type to match the source task to infer the recognition information to the task of the source domain. For example, if $T_t$ is an action recognition problem using 3D-CNN, the input $x^{w\times h\times d}\in D_t$ is defined as a three-dimensional tensor. In this case, if a pretrained network for transfer learning is obtained through the image recognition problem $T_s$ using 2D-CNN, $f_s:x^{w\times h\times d}\to\hat{x}^{w\times h}$ should be defined as a function that maps a three-dimensional tensor to a two-dimensional matrix into which $h_s$ can be input. Up to $M$ number of different type of transformation functions could be defined. Algorithm1 describes how the SSKT works depending on each transfer learning scenario.

\smallskip \noindent \textbf{Transfer Modules Depending on CNN Architecture.} To encourage predicting $y_{s,i}^{aux}$ by $h_t$, we design bottleneck structure based transfer module supporting auxiliary task using feature output from each convolutional block. Figure 1 shows configuration of transfer modules depending on each CNN architecture for its problem domain. We applied the transfer module to four different CNN architectures such as ResNet \cite{He16}, DenseNet \cite{Huang17}, MobileNet \cite{Sandler18}, and 3D-ResNet \cite{Hara18} for each problem domain. 

\begin{figure}[t!]
\centering
\includegraphics[width=1.0\linewidth]{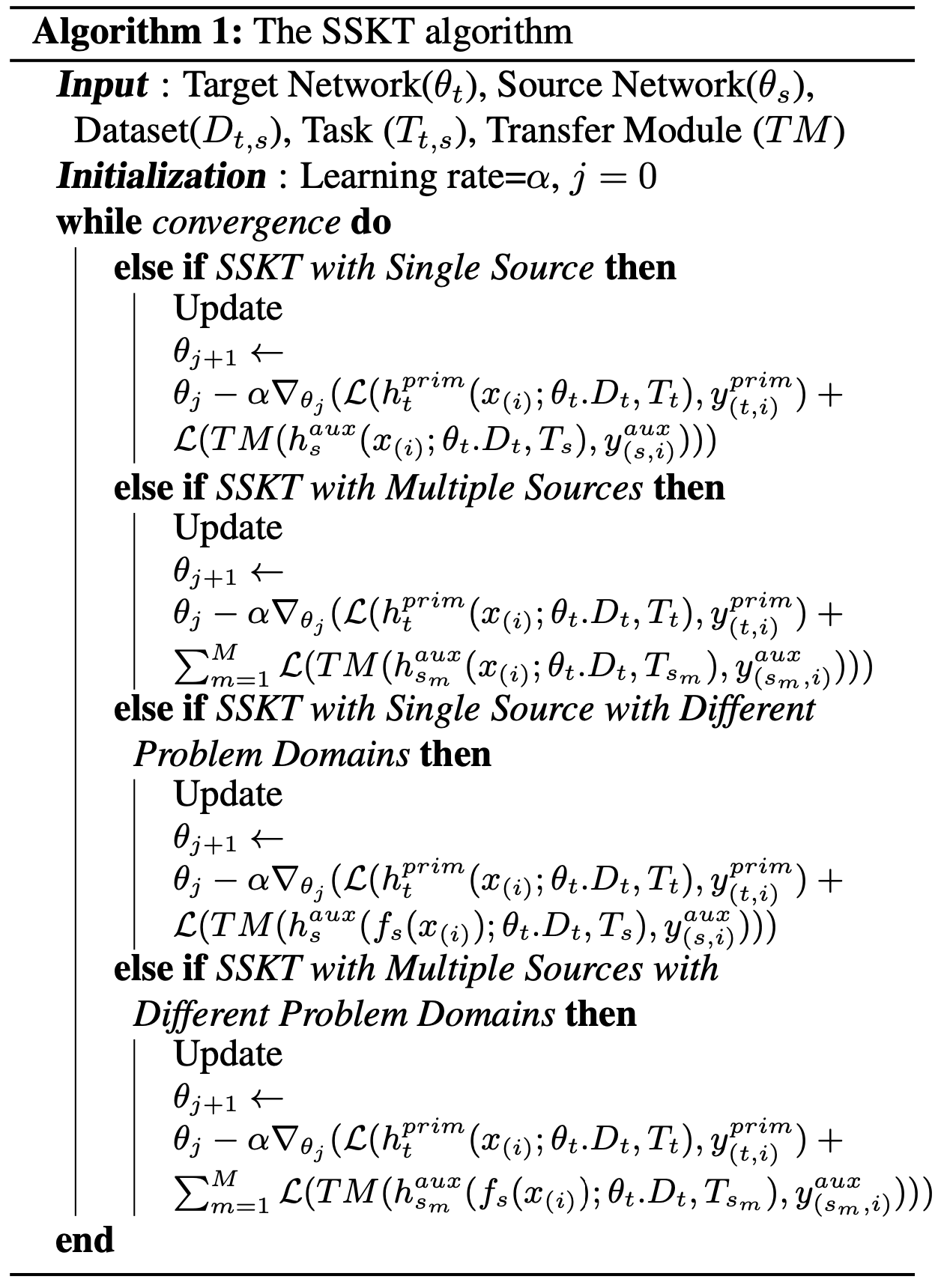}
\label{fig:alg}
\end{figure}

\begin{figure*}[t!]
\centering
\includegraphics[width=1.0\linewidth]{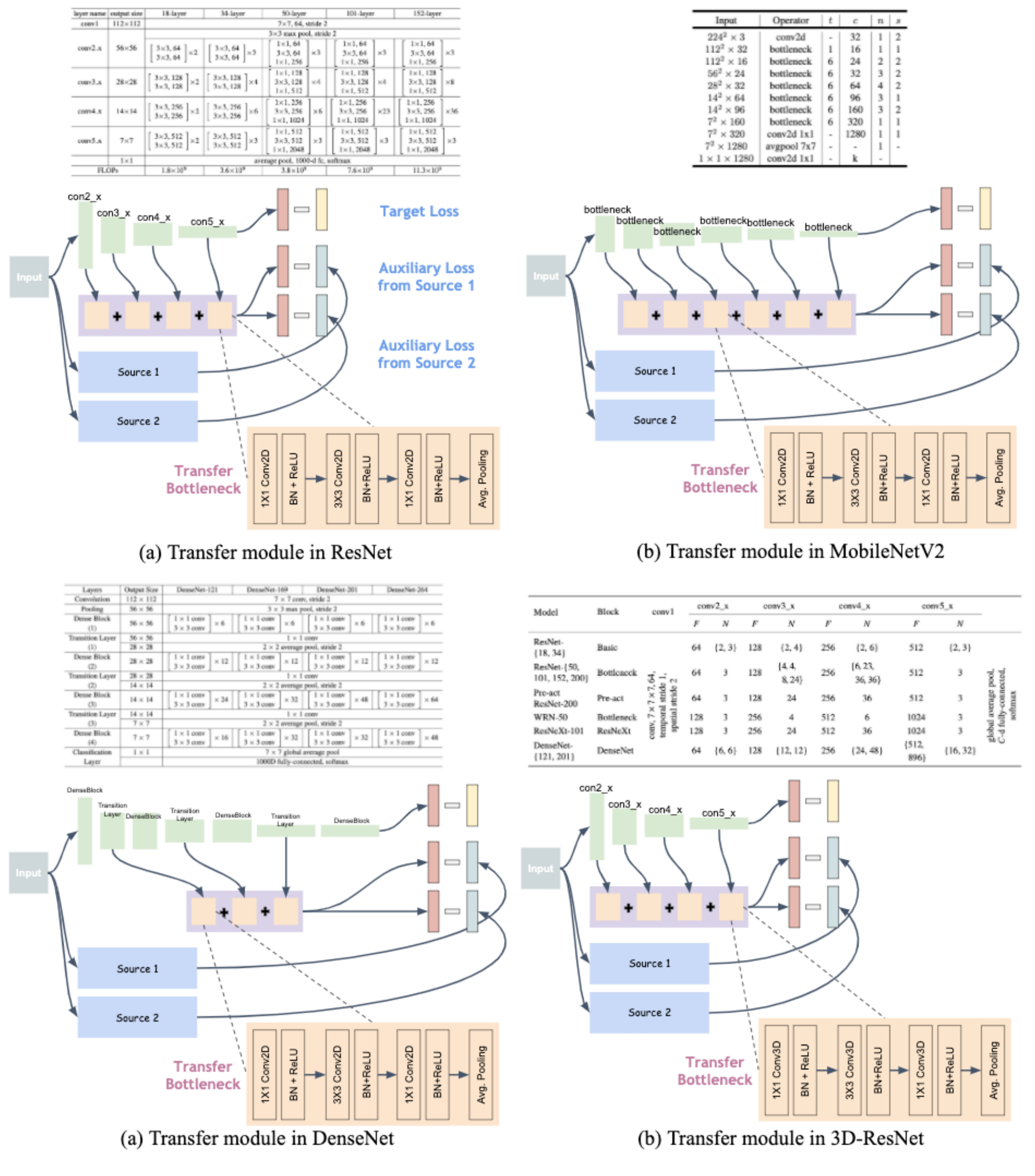}
\caption{\textbf{Schematic of the transfer modules for efficient SSKT with different CNN architectures.} The transfer module used in the SSKT consists of summation of feature output of bottleneck layers from each convolutional block. Schematic shows and example of the transfer module with different CNN architectures for SSKT using multiple sources.}
\label{fig:example}
\end{figure*}

\section*{B. Additional Experiment Results}
We provide performance for all experimental configuraltions for each dataset, in addition to the results contained in the main manuscript. For fair comparison of SSKT, the configurations consist of a combination of the type of source and target  network, the presence or absence of a transfer module, and a loss function. For model architecture and hyperparameters setting for training (See Table 1 of the main manuscript).  Same as the experiment results of the main manuscript, the datasets of the source task are ImageNet \cite{Deng09} and Places365 \cite{Zhou18}, and the datasets of the target task are CIFAR10/100 \cite{Krizhevsky09}, STL10 \cite{Coates11}, ImageNet, Places365, PASCAL VOC \cite{Everingham15}, UCF101 \cite{Soomro12}, and HMDB51 \cite{Kuehne11}. Tables 1 to 6 provide performance according to the experimental conditions of each dataset. Figure 2 shows the performance changes for  the STL10 and PASCAL VOC datasets depending on hyperparameters, the structure of the source and target network, and the presence or absence of a transfer module. The abbreviations for the datasets and model architectures listed in all experimental tables are as follows: 

\smallskip \noindent \textbf{Datasets}: ImageNet (I), Places365 (P), CIFAR10 (C10), CIFAR100 (C100), STL10 (S10), PASCAL VOC (VOC), UCF101 (U101), and HMDB51 (H51).

\smallskip \noindent \textbf{Model architectures}: ResNet (R), DenseNet (D), MobileNetV2 (MV2), and 3D-ResNet (3DR).

\smallskip Finally, we included the experimental results according to the training setting for further analysis of the SSKT. Table 7 shows the all the combination of comparison results for DenseNet121 and MoblieNetV2, and Table 8 shows the evaluations for each experimental setting with finetuning scenario. Table 9 shows the performance comparison with MAXL \cite{Liu19} which is the state-of-the-art self-supervised learning based on auxiliary learning.

\begin{table}[t!]
  \caption{Performance change according to the configuration of the SSKT for CIFAR10 dataset compared to the training from scratch. All experiments evaluated test performance 3 times from the same random seed for the model. TM stands for Transfer Module and R[depth] stands for ResNet structure. The best performance of each network architecture highlighted in \textbf{bold}.}
  \label{tab:freq}
  \resizebox{\linewidth}{!}{
  \begin{tabular}{c|c|c|c|c|c|c}
%    \toprule
\hline
   $T_s$ 	& 	$T_t$ 	& Model & Method & TM & Loss & acc.\\ \hline
 %   \midrule 
 	- 	& C10 		& R20 & scratch & - & CE & 92.19$\pm$0.09\\ \hline
P 		&  			& R20 & SSKT & x & CE+CE & 92.21$\pm$0.06\\ 
		&  			& R20 & SSKT & x & CE+KD & 92.24$\pm$0.14\\ 
 		&  			& R20 & SSKT & o & CE+CE & 92.23$\pm$0.04\\ 
		&  			& R20 & SSKT & o & CE+KD & 92.25$\pm$0.04\\ \hline
I 		&  			& R20 & SSKT & x & CE+CE & 92.28$\pm$0.07\\ 
 		&  			& R20 & SSKT & x & CE+KD & 92.34$\pm$0.07\\ 
		&  			& R20 & SSKT & o & CE+CE & 92.44$\pm$0.05\\ 
		&  			& R20 & SSKT & o & CE+KD & 92.29$\pm$0.0\\ \hline
P+I		& 			& R20 & SSKT & x & CE+CE & 91.9$\pm$0.1\\ 
		&  			& R20 & SSKT & x & CE+KD & \textbf{92.46$\pm$0.15}\\ 
		&  			& R20 & SSKT & o & CE+CE & 92.42$\pm$0.07\\ 
		&  			& R20 & SSKT & o & CE+KD & 92.22$\pm$0.17\\ \hline \hline
 	- 	& 	 		& R32 & scratch & - & CE & 93.21$\pm$0.09\\ \hline
P 		&  			& R32 & SSKT & x & CE+CE & 92.77$\pm$0.14\\ 
		&  			& R32 & SSKT & x & CE+KD & 92.87$\pm$0.31\\ 
 		&  			& R32 & SSKT & o & CE+CE & 92.65$\pm$0.26\\ 
		&  			& R32 & SSKT & o & CE+KD & 92.59$\pm$0.22\\ \hline
I 		&  			& R32 & SSKT & x & CE+CE & 93.26$\pm$0.08\\ 
 		&  			& R32 & SSKT & x & CE+KD & 92.78$\pm$0.2\\ 
		&  			& R32 & SSKT & o & CE+CE & 93.25$\pm$0.12\\ 
		&  			& R32 & SSKT & o & CE+KD & 92.88$\pm$0.07\\ \hline
P+I		& 			& R32 & SSKT & x & CE+CE & 92.88$\pm$0.15\\ 
		&  			& R32 & SSKT & x & CE+KD & 93.07$\pm$0.09\\ 
		&  			& R32 & SSKT & o & CE+CE & \textbf{93.38$\pm$0.02}\\ 
		&  			& R32 & SSKT & o & CE+KD & 93.1$\pm$0.22\\ \hline
\end{tabular}
}
\end{table}

\begin{table}[t!]
  \caption{Performance change according to the configuration of the SSKT for CIFAR100 dataset  compared to the training from scratch. }
  \label{tab:freq}
  \resizebox{\linewidth}{!}{
  \begin{tabular}{c|c|c|c|c|c|c}
%    \toprule
\hline
   $T_s$ 	& 	$T_t$ 	& Model & Method & TM & Loss & acc.\\ \hline
 %   \midrule 
 	- 	& C100		& R20 & scratch & - & CE & 68.26$\pm$0.36\\ \hline
P 		&  			& R20 & SSKT & x & CE+CE & 67.65$\pm$0.21\\ 
		&  			& R20 & SSKT & x & CE+KD & 68.01$\pm$0.42\\ 
 		&  			& R20 & SSKT & o & CE+CE & 67.96$\pm$0.27\\ 
		&  			& R20 & SSKT & o & CE+KD & 67.83$\pm$0.34\\ \hline
I 		&  			& R20 & SSKT & x & CE+CE & 68.3$\pm$0.17\\ 
 		&  			& R20 & SSKT & x & CE+KD & 68.37$\pm$0.23\\ 
		&  			& R20 & SSKT & o & CE+CE & \textbf{68.63$\pm$0.12}\\ 
		&  			& R20 & SSKT & o & CE+KD & 68.35$\pm$0.1\\ \hline
P+I		& 			& R20 & SSKT & x & CE+CE & 67.87$\pm$0.17\\ 
		&  			& R20 & SSKT & x & CE+KD & 68.13$\pm$0.05\\ 
		&  			& R20 & SSKT & o & CE+CE & 68.56$\pm$0.23\\ 
		&  			& R20 & SSKT & o & CE+KD & 67.84$\pm$0.28\\ \hline \hline
 	- 	& 	 		& R32 & scratch & - & CE & 70.33$\pm$0.19\\ \hline
P 		&  			& R32 & SSKT & x & CE+CE & 69.97$\pm$0.16\\ 
		&  			& R32 & SSKT & x & CE+KD & 69.93$\pm$0.21\\ 
 		&  			& R32 & SSKT & o & CE+CE & 69.69$\pm$0.19\\ 
		&  			& R32 & SSKT & o & CE+KD & 69.92$\pm$0.31\\ \hline
I 		&  			& R32 & SSKT & x & CE+CE & 70.6$\pm$0.05\\ 
 		&  			& R32 & SSKT & x & CE+KD & 70.17$\pm$0.14\\ 
		&  			& R32 & SSKT & o & CE+CE & 70.75$\pm$0.06\\ 
		&  			& R32 & SSKT & o & CE+KD & 70.0$\pm$0.11\\ \hline
P+I		& 			& R32 & SSKT & x & CE+CE & 69.25$\pm$0.58\\ 
		&  			& R32 & SSKT & x & CE+KD & 69.22$\pm$0.43\\ 
		&  			& R32 & SSKT & o & CE+CE & \textbf{70.94$\pm$0.36}\\ 
		&  			& R32 & SSKT & o & CE+KD &  69.44$\pm$0.01\\ \hline
\end{tabular}
}
\end{table}

\begin{table}[t!]
  \caption{Performance change according to the configuration of the SSKT for STL10 dataset  compared to the training from scratch. }
  \label{tab:freq}
  \resizebox{\linewidth}{!}{
  \begin{tabular}{c|c|c|c|c|c|c}
%    \toprule
\hline
   $T_s$ 	& 	$T_t$ 	& Model & Method & TM & Loss & acc.\\ \hline
 %   \midrule 
 	- 	& STL10		& R20 & scratch & - & CE & 81.15$\pm$0.34\\ \hline
P 		&  			& R20 & SSKT & x & CE+CE & 81.56$\pm$0.32\\ 
		&  			& R20 & SSKT & x & CE+KD & 80.88$\pm$0.19\\ 
 		&  			& R20 & SSKT & o & CE+CE & 82.76$\pm$0.05\\ 
		&  			& R20 & SSKT & o & CE+KD & 81.06$\pm$0.2\\ \hline
I 		&  			& R20 & SSKT & x & CE+CE & 82.2$\pm$0.17\\ 
 		&  			& R20 & SSKT & x & CE+KD & 80.82$\pm$0.14\\ 
		&  			& R20 & SSKT & o & CE+CE & 83.45$\pm$0.07\\ 
		&  			& R20 & SSKT & o & CE+KD & 81.3$\pm$0.39\\ \hline
P+I		& 			& R20 & SSKT & x & CE+CE & 82.46$\pm$0.24\\ 
		&  			& R20 & SSKT & x & CE+KD & 81.47$\pm$0.22\\ 
		&  			& R20 & SSKT & o & CE+CE & \textbf{84.56$\pm$0.35}\\ 
		&  			& R20 & SSKT & o & CE+KD & 81.33$\pm$0.11\\ \hline \hline
 	- 	& 	 		& R32 & scratch & - & CE & 81.19$\pm$0.17\\ \hline
P 		&  			& R32 & SSKT & x & CE+CE & 82.1$\pm$0.14\\ 
		&  			& R32 & SSKT & x & CE+KD & 81.29$\pm$0.22\\ 
 		&  			& R32 & SSKT & o & CE+CE & 83.06$\pm$0.27\\ 
		&  			& R32 & SSKT & o & CE+KD & 81.19$\pm$0.12\\ \hline
I 		&  			& R32 & SSKT & x & CE+CE & 82.88$\pm$0.33\\ 
 		&  			& R32 & SSKT & x & CE+KD & 81.4$\pm$0.23\\ 
		&  			& R32 & SSKT & o & CE+CE & 83.68$\pm$0.28\\ 
		&  			& R32 & SSKT & o & CE+KD & 81.76$\pm$0.18\\ \hline
P+I		& 			& R32 & SSKT & x & CE+CE & 82.39$\pm$0.15\\ 
		&  			& R32 & SSKT & x & CE+KD & 79.8$\pm$0.47\\ 
		&  			& R32 & SSKT & o & CE+CE & \textbf{83.4$\pm$0.2}\\ 
		&  			& R32 & SSKT & o & CE+KD &  80.05$\pm$1.06\\ \hline
\end{tabular}
}
\end{table}

\begin{table}[t!]
  \caption{Performance change according to the configuration of the SSKT for ImageNet and Places365 compared to the training from scratch.}
  \label{tab:freq}
  \resizebox{\linewidth}{!}{
  \begin{tabular}{c|c|c|c|c|c|c}
%    \toprule
\hline
   $T_s$ 	& 	$T_t$ 	& Model & Method & TM & Loss & acc.\\ \hline
 %   \midrule 
 	- 	& P	 		& R18 & scratch & - & CE & 50.92\\ \hline
P 		&  			& R18 & SSKT & x & CE+CE & 54.41\\ 
		&  			& R18 & SSKT & x & CE+KD & 53.42\\ 
		&  			& R18 & SSKT & o & CE+CE & 54.5\\ 
		&  			& R18 & SSKT & o & CE+KD & 54.11\\ \hline 
I 		&  			& R18 & SSKT & x & CE+CE & 53.47\\ 
		&  			& R18 & SSKT & x & CE+KD & 53.51\\ 
		&  			& R18 & SSKT & o & CE+CE & 53.67\\
		&  			& R18 & SSKT & o & CE+KD & 53.44\\ \hline
P+I		&  			& R18 & SSKT & x & CE+CE & \textbf{54.78}\\ 
		&  			& R18 & SSKT & x & CE+KD & 54.5\\ 
		&  			& R18 & SSKT & o & CE+CE & 54.62\\ 
		&  			& R18 & SSKT & o & CE+KD & 54.5\\ \hline \hline 
 	- 	& I	 		& R18 & scratch & - & CE & 64.14\\ \hline
P 		&  			& R18 & SSKT & x & CE+CE & 64.18\\
		&  			& R18 & SSKT & x & CE+KD & 64.21\\ 
		&  			& R18 & SSKT & o & CE+CE & 64.99\\ 
		&  			& R18 & SSKT & o & CE+KD & 63.53\\ \hline \hline 
I 		&  			& R18 & SSKT & x & CE+CE & 67.79\\ 
		&  			& R18 & SSKT & x & CE+KD & 66.0\\ 
		&  			& R18 & SSKT & o & CE+CE & 67.46\\ 
		&  			& R18 & SSKT & o & CE+KD & 65.65\\ \hline \hline
P+I		&  			& R18 & SSKT & x & CE+CE & \textbf{70.57}\\ 
		&  			& R18 & SSKT & x & CE+KD & 67.42\\ 
		&  			& R18 & SSKT & o & CE+CE & 67.64\\ 
		&  			& R18 & SSKT & o & CE+KD & 66.81\\ \hline
\end{tabular}
}
\end{table}

\begin{table}[t!]
  \caption{Performance change according to the configuration of the SSKT for PASCAL VOC compared to the training from scratch.}
  \label{tab:freq}
  \resizebox{\linewidth}{!}{
  \begin{tabular}{c|c|c|c|c|c|c}
%    \toprule
\hline
   $T_s$ 	& 	$T_t$ 	& Model & Method & TM & Loss & acc. \\ \hline
 %   \midrule 
 	- 	& VOC	 	& R18 & scratch & - & BCE & 67.28$\pm$0.25\\ \hline
P 		&  			& R18 & SSKT & x & BCE+CE & 74.34$\pm$0.23\\ 
 		&  			& R18 & SSKT & x & BCE+KD & 69.92$\pm$0.16\\ 
 		&  			& R18 & SSKT & o & BCE+CE & 74.76$\pm$0.17\\ 
 		&  			& R18 & SSKT & o & BCE+KD & 69.9$\pm$0.18\\ \hline
I 		&  			& R18 & SSKT & x & BCE+CE & 74.78$\pm$0.09\\ 
 		&  			& R18 & SSKT & x & BCE+KD & 69.9$\pm$0.35\\ 
 		&  			& R18 & SSKT & o & BCE+CE & 74.58$\pm$0.11\\ 
 		&  			& R18 & SSKT & o & BCE+KD & 69.95$\pm$0.19\\ \hline
P+I		&  			& R18 & SSKT & x & BCE+CE & 76.33$\pm$0.0\\ 
 		&  			& R18 & SSKT & x & BCE+KD & 69.74$\pm$0.17\\ 
 		&  			& R18 & SSKT & o & BCE+CE & \textbf{76.42$\pm$0.06}\\ 
 		&  			& R18 & SSKT & o & BCE+KD & 69.89$\pm$0.13\\ \hline \hline
 	- 	& 		 	& R34 & scratch & - & BCE & 66.0$\pm$0.49\\ \hline
P 		&  			& R34 & SSKT & x & BCE+CE & 73.83$\pm$0.38\\ 
 		&  			& R34 & SSKT & x & BCE+KD & 69.93$\pm$0.03\\ 
 		&  			& R34 & SSKT & o & BCE+CE & 75.65$\pm$0.12\\ 
 		&  			& R34 & SSKT & o & BCE+KD & 69.51$\pm$0.13\\ \hline
I 		&  			& R34 & SSKT & x & BCE+CE & 74.25$\pm$0.12\\ 
 		&  			& R34 & SSKT & x & BCE+KD & 70.05$\pm$0.14\\ 
 		&  			& R34 & SSKT & o & BCE+CE & 75.14$\pm$0.14\\ 
 		&  			& R34 & SSKT & o & BCE+KD & 70.18$\pm$0.11\\ \hline
P+I		&  			& R34 & SSKT & x & BCE+CE & 75.88$\pm$0.1\\ 
 		&  			& R34 & SSKT & x & BCE+KD & 70.15$\pm$0.09\\ 
 		&  			& R34 & SSKT & o & BCE+CE & \textbf{77.02$\pm$0.02}\\ 
 		&  			& R34 & SSKT & o & BCE+KD & 70.58$\pm$0.35\\ \hline \hline
 	- 	& 		 	& R50 & scratch & - & BCE & 61.16$\pm$0.34\\ \hline
P 		&  			& R50 & SSKT & x & BCE+CE & 63.29$\pm$1.43\\ 
 		&  			& R50 & SSKT & x & BCE+KD & 65.5$\pm$0.2\\ 
 		&  			& R50 & SSKT & o & BCE+CE & 74.44$\pm$0.06\\ 
 		&  			& R50 & SSKT & o & BCE+KD & 65.94$\pm$0.09\\ \hline
I 		&  			& R50 & SSKT & x & BCE+CE & 63.96$\pm$2.74\\ 
 		&  			& R50 & SSKT & x & BCE+KD & 66.11$\pm$0.32\\ 
 		&  			& R50 & SSKT & o & BCE+CE & 74.24$\pm$0.05\\ 
 		&  			& R50 & SSKT & o & BCE+KD & 65.77$\pm$0.13\\ \hline
P+I		&  			& R50 & SSKT & x & BCE+CE & 69.27$\pm$0.21\\ 
 		&  			& R50 & SSKT & x & BCE+KD & 66.0$\pm$0.29\\ 
 		&  			& R50 & SSKT & o & BCE+CE & \textbf{77.1$\pm$0.14}\\ 
 		&  			& R50 & SSKT & o & BCE+KD & 66.22$\pm$0.23\\ \hline
\end{tabular}
}
\end{table}

\begin{table}[t!]
  \caption{Performance change according to the configuration of the SSKT for UCF101 and HMDB51 compared to the training from scratch.}
  \label{tab:freq}
  \resizebox{\linewidth}{!}{
  \begin{tabular}{c|c|c|c|c|c|c}
%    \toprule
\hline
   $T_s$ 	& 	$T_t$ 	& Model & Method & TM & Loss & acc.\\ \hline
 %   \midrule 
 	- 	& U101	 	& 3DR18 & scratch & - & CE & 43.28\\ \hline
P 		&  			& 3DR18 & SSKT & x & CE+CE & 44.1\\ 
		&  			& 3DR18 & SSKT & x & CE+KD & 44.79\\ 
		&  			& 3DR18 & SSKT & o & CE+CE & 45.35\\ 
		&  			& 3DR18 & SSKT & o & CE+KD & 43.95\\ \hline
I 		&  			& 3DR18 & SSKT & x & CE+CE & 46.62\\ 
		&  			& 3DR18 & SSKT & x & CE+KD & 40.35\\ 
		&  			& 3DR18 & SSKT & o & CE+CE & 44.26\\ 
		&  			& 3DR18 & SSKT & o & CE+KD & 38.95\\ \hline
P+I		&  			& 3DR18 & SSKT & x & CE+CE & \textbf{52.19}\\ 
		&  			& 3DR18 & SSKT & x & CE+KD & 43.68\\ 
		&  			& 3DR18 & SSKT & o & CE+CE & 47.09\\ 
		&  			& 3DR18 & SSKT & o & CE+KD & 45.0\\ \hline \hline
 	- 	& H51	 	& 3DR18 & scratch & - & CE & 17.14\\ \hline
P 		&  			& 3DR18 & SSKT & x & CE+CE & 18.18\\ 
		&  			& 3DR18 & SSKT & x & CE+KD & 17.33\\ 
		&  			& 3DR18 & SSKT & o & CE+CE & 18.77\\ 
		&  			& 3DR18 & SSKT & o & CE+KD & 17.59\\ \hline
I 		&  			& 3DR18 & SSKT & x & CE+CE & 18.64\\ 
		&  			& 3DR18 & SSKT & x & CE+KD & 18.12\\ 
		&  			& 3DR18 & SSKT & o & CE+CE & 18.38\\ 
		&  			& 3DR18 & SSKT & o & CE+KD & 18.77\\ \hline
P+I		&  			& 3DR18 & SSKT & x & CE+CE & 19.75\\ 
		&  			& 3DR18 & SSKT & x & CE+KD & 18.38\\ 
		&  			& 3DR18 & SSKT & o & CE+CE & \textbf{20.54}\\ 
		&  			& 3DR18 & SSKT & o & CE+KD & 17.99\\ \hline
\end{tabular}
}
\end{table}

 \begin{figure*}[t!]
\centering
\includegraphics[width=1.0\linewidth]{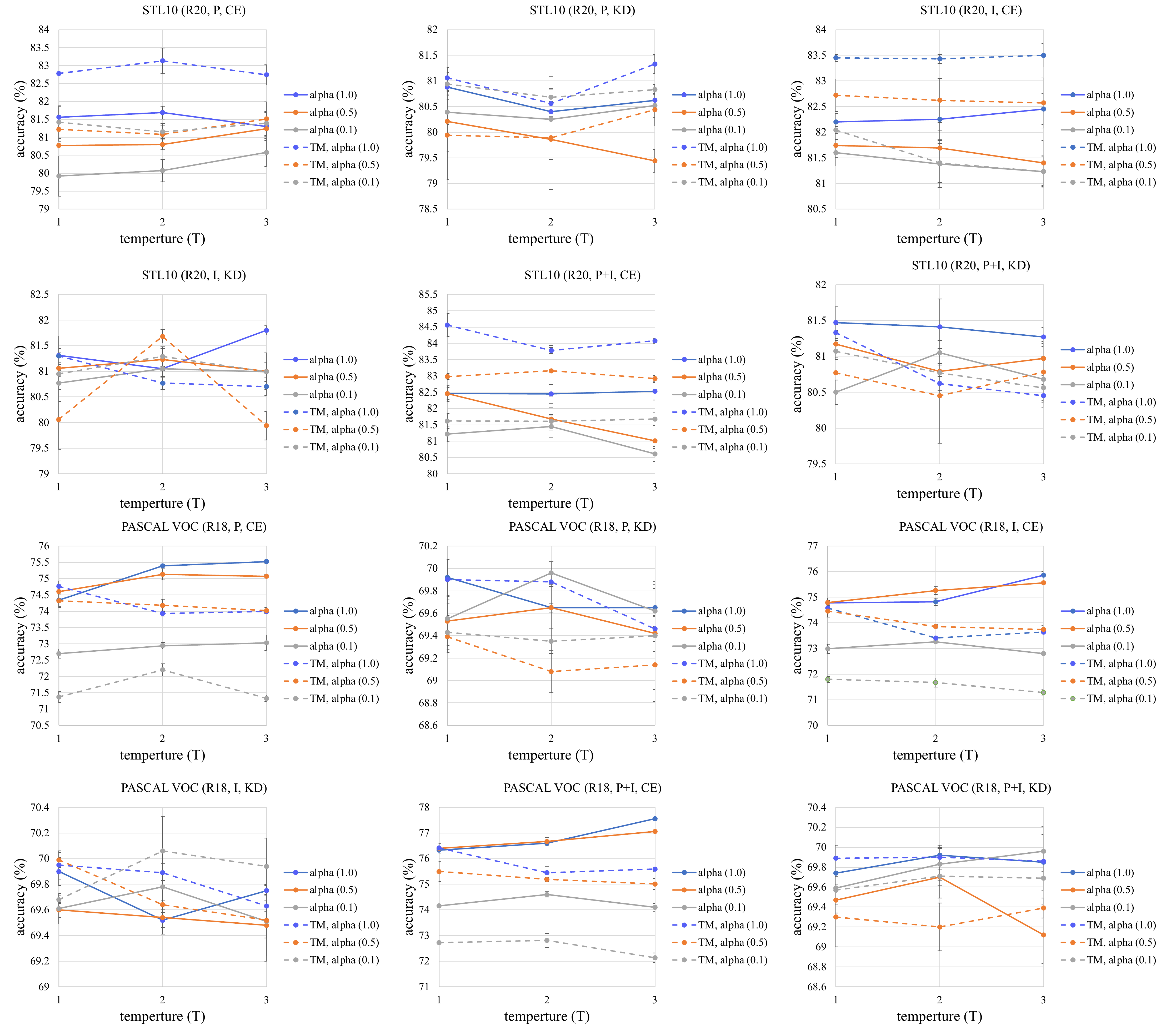}
\caption{\textbf{Parameter optimization of SSKT.} The title of each graph is composed of  $D_t$ (target model, $T_s$, auxiliary loss). $T$ is the temparture parameter of each auxiliary loss, and $\alpha$ is the balance parameter of the total loss.}
\label{fig:example}
\end{figure*}

\begin{table}[t!]
  \caption{Performance change according to the configuration of the SSKT for STL10 compared to the training from scratch with MobileNet V2 (MV2) and DenseNet121 (D121).}
  \label{tab:freq}
  \resizebox{\linewidth}{!}{
  \begin{tabular}{c|c|c|c|c|c|c}
%    \toprule
\hline
   $T_s$ 	& 	$T_t$ 	& Model & Method & TM & Loss & acc.\\ \hline
 %   \midrule 
 	- 	& S10		& MV2 & scratch & - & CE & 72.26$\pm$0.83\\ \hline 
P		&  			& MV2 & SSKT & x & CE+CE & 75.79$\pm$0.19\\ 
		&  			& MV2 & SSKT & x & CE+KD & 74.0$\pm$0.35\\ 
		&  			& MV2 & SSKT & o & CE+CE & 75.28$\pm$0.49\\ 
		&  			& MV2 & SSKT & o & CE+KD & 73.37$\pm$1.8\\ \hline
I		&  			& MV2 & SSKT & x & CE+CE & 76.08$\pm$0.63\\ 
		&  			& MV2 & SSKT & x & CE+KD & 74.39$\pm$0.82\\ 
		&  			& MV2 & SSKT & o & CE+CE & 75.35$\pm$0.61\\ 
		&  			& MV2 & SSKT & o & CE+KD & 72.6$\pm$0.67\\ \hline
P+I		&  			& MV2 & SSKT & x & CE+CE & 76.69$\pm$0.18\\ 
		&  			& MV2 & SSKT & x & CE+KD & 73.29$\pm$0.89\\ 
		&  			& MV2 & SSKT & o & CE+CE & \textbf{76.96$\pm$0.39}\\ 
		&  			& MV2 & SSKT & o & CE+KD & 73.35$\pm$0.99\\ \hline \hline
 	- 	& 		 	& D121 & scratch & - & CE & 72.02$\pm$0.48\\ \hline
P		& 			& D121 & SSKT & x & CE+CE & 76.17$\pm$0.35\\
		& 			& D121 & SSKT & x & CE+KD & 74.83$\pm$0.59\\
		& 			& D121 & SSKT & o & CE+CE & 73.46$\pm$0.62\\
		& 			& D121 & SSKT & o & CE+KD & 72.55$\pm$0.43\\ \hline
I		& 			& D121 & SSKT & x & CE+CE & 76.0$\pm$0.33\\
		& 			& D121 & SSKT & x & CE+KD & 73.7$\pm$0.22\\
		& 			& D121 & SSKT & o & CE+CE & 74.35$\pm$0.3\\
		& 			& D121 & SSKT & o & CE+KD & 71.13$\pm$0.59\\ \hline
P+I		& 			& D121 & SSKT & x & CE+CE & \textbf{77.03$\pm$0.17}\\
		& 			& D121 & SSKT & x & CE+KD & 73.76$\pm$0.84\\
		& 			& D121 & SSKT & o & CE+CE & 76.09$\pm$0.26\\
		& 			& D121 & SSKT & o & CE+KD & 70.94$\pm$1.14\\ \hline
\end{tabular}
}
\end{table}

\begin{table}[t!]
  \caption{SSKT results for PASCAL VOC, UCF101, and HMDB51 using pretrained weights. \textit{ft} stands for finetuning and K stands for Kinetics-400 dataset (Kay et al. 2017).}
  \label{tab:freq}
  \resizebox{\linewidth}{!}{
  \begin{tabular}{c|c|c|c|c|c|c}
%    \toprule
\hline
   $T_s$ 	& 	$T_t$ 	& Model & Method & TM & Loss & acc.\\ \hline
 %   \midrule 
 	- 	& VOC	 	& R18 & \textit{ft} (I) & - & CE & 90.52$\pm$0.11\\ \hline
P 		&  			& R18 & SSKT & x & CE+CE & 89.3$\pm$0.04\\ 
 		&  			& R18 & SSKT & x & CE+KD & \textbf{92.28$\pm$0.06}\\ 
 		&  			& R18 & SSKT & o & CE+CE & 90.83$\pm$0.04\\ 
 		&  			& R18 & SSKT & o & CE+KD & 92.21$\pm$0.05\\ \hline
I 		&  			& R18 & SSKT & x & CE+CE & 91.29$\pm$0.03\\ 
		&  			& R18 & SSKT & x & CE+KD & 92.26$\pm$0.07\\ 
		&  			& R18 & SSKT & o & CE+CE & 91.58$\pm$0.15\\ 
		&  			& R18 & SSKT & o & CE+KD & 92.19$\pm$0.09\\ \hline
P+I		& 			& R18 & SSKT & x & CE+CE & 91.28$\pm$0.05\\ 
		&  			& R18 & SSKT & x & CE+KD & 92.19$\pm$0.07\\ 
		&  			& R18 & SSKT & o & CE+CE & 91.25$\pm$0.08\\ 
		&  			& R18 & SSKT & o & CE+KD & 92.25$\pm$0.07\\ \hline \hline
 	- 	& U101	 	& 3DR18 & \textit{ft} (K) & - & CE & 83.95\\ \hline
P 		&  			& 3DR18 & SSKT & x & CE+CE & 84.53\\ 
 		&  			& 3DR18 & SSKT & x & CE+KD & \textbf{84.58}\\ 
 		&  			& 3DR18 & SSKT & o & CE+CE & 83.87\\ 
 		&  			& 3DR18 & SSKT & o & CE+KD & 83.98\\ \hline
I 		&  			& 3DR18 & SSKT & x & CE+CE & 81.99\\ 
 		&  			& 3DR18 & SSKT & x & CE+KD & 83.42\\ 
 		&  			& 3DR18 & SSKT & o & CE+CE & 84.29\\ 
 		&  			& 3DR18 & SSKT & o & CE+KD & 84.37\\ \hline
P+I		& 			& 3DR18 & SSKT & x & CE+CE & 78.56\\ 
		& 			& 3DR18 & SSKT & x & CE+KD & 84.14\\ 
		& 			& 3DR18 & SSKT & o & CE+CE & 82.81\\
		& 			& 3DR18 & SSKT & o & CE+KD & 84.19\\\hline \hline
 	- 	& H51	 	& 3DR18 & \textit{ft} (K) & - & CE & 56.64\\ \hline
P 		&  			& 3DR18 & SSKT & x & CE+CE & 56.77\\ 
 		&  			& 3DR18 & SSKT & x & CE+KD & 56.77\\ 
 		&  			& 3DR18 & SSKT & o & CE+CE & 57.75\\ 
 		&  			& 3DR18 & SSKT & o & CE+KD & \textbf{57.82}\\ \hline
I 		&  			& 3DR18 & SSKT & x & CE+CE & 56.18\\ 
 		&  			& 3DR18 & SSKT & x & CE+KD & 56.9\\ 
 		&  			& 3DR18 & SSKT & o & CE+CE & 53.3\\ 
 		&  			& 3DR18 & SSKT & o & CE+KD & 57.75\\ \hline
P+I		& 			& 3DR18 & SSKT & x & CE+CE & 54.48\\ 
		& 			& 3DR18 & SSKT & x & CE+KD & 56.05\\ 
		& 			& 3DR18 & SSKT & o & CE+CE & 57.29\\ 
		& 			& 3DR18 & SSKT & o & CE+KD & 57.1\\ \hline
\end{tabular}
}
\end{table}

\begin{table}[t!]
  \caption{Comparison with MAXL \cite{Liu19} according to the configuration of the SSKT. V16 denotes VGG16 \cite{Simonyan15}. F denotes focal loss \cite{Liu19}. C denotes cross-entropy loss. $TM$ denotes transfer module.}.
  \label{tab:freq}
  \resizebox{\linewidth}{!}{
  \begin{tabular}{c|c|c|c|c}
%    \toprule
\hline
$T_t$  & $T_t$ Model & MAXL ($\psi[i]$) & SSKT ($T_s$, $TM$) & $T_s$ Model  \\ \hline
 %   \midrule 
C10 		& V16 (F) & 93.27$\pm$0.09 (2) &  93.56$\pm$0.02 (I, x) & V16 \\ 
		& V16 (F) & 93.47$\pm$0.08 (3) &  94.1$\pm$0.1 (I, o) & V16 \\
		& V16 (F) & 93.49$\pm$0.05 (5) &  92.94$\pm$0.02 (I, x) & R50 \\ 
		& V16 (F) & 93.10$\pm$0.08 (10) &  94.1$\pm$0.1 (I, o) & R50 \\  
		& V16 (F) & - & 92.56$\pm$0.15 (P, x) & R50 \\ 
		& V16 (F) & - & 92.72$\pm$0.15 (P, o) & R50 \\ 	
		& V16 (F) & - & 92.6$\pm$0.19 (P+I, x) & R50, V16 \\
		& V16 (F) & - & 93.06$\pm$0.2  (P+I, o) & R50, V16 \\ 
		& V16 (F) & - &  92.2$\pm$0.08 (P+I, x) & R50, R50 \\
		& V16 (F) & - & 92.54$\pm$0.17 (P+I, o) & R50, R50 \\ 
		& V16 (C) & - & 93.78$\pm$0.04 (I, x) & V16 \\ 
		& V16 (C) & - & \textbf{94.22$\pm$0.02} (I, o) & V16 \\ 
		& V16 (C) & - & 93.08$\pm$0.06  (I, x) & R50 \\
		& V16 (C) & - & 93.78$\pm$0.04 (I, o) & R50 \\
		& V16 (C) & - & 93.04$\pm$0.15 (P, x) & R50 \\  
		& V16 (C) & - & 93.12$\pm$0.12 (P, o) & R50 \\  
		& V16 (C) & - & 93.35$\pm$0.21 (P+I, x) & R50, V16 \\
		& V16 (C) & - & 93.67$\pm$0.17 (P+I, o) & R50, V16 \\
		& V16 (C) & - & 93.02$\pm$0.11 (P+I, x) & R50, R50 \\ 
		& V16 (C) & - & 93.26$\pm$0.12 (P+I, o) & R50, R50 \\ \hline
		& R20 (F) & 91.53$\pm$0.33 (2) &  90.52$\pm$0.34 (I, x) & V16 \\ 
		& R20 (F) & 91.52$\pm$0.1 (3) &  91.48$\pm$0.03 (I, o) & V16 \\ 
		& R20 (F) & 91.38$\pm$0.47 (5) &  90.88$\pm$0.02 (I, x) & R50 \\ 
		& R20 (F) & 91.56$\pm$0.16 (10) &  91.66$\pm$0.09 (I, o) & R50 \\ 
		& R20 (F) & - &  89.97$\pm$0.02 (P, x) & R50 \\ 
		& R20 (F) & - &  91.26$\pm$0.17 (P, o) & R50 \\ 
		& R20 (F) & - &  89.42$\pm$0.08 (P+I, x) & R50, V16 \\ 		
		& R20 (F) & - &  90.93$\pm$0.01 (P+I, o) & R50, V16 \\ 
		& R20 (F) & - &  90.02$\pm$0.21 (P+I, x) & R50, R50 \\
		& R20 (F) & - &  91.11$\pm$0.18 (P+I, o) & R50, R50 \\ 
		& R20 (C) & - &  92.28$\pm$0.07 (I, x) & R50 \\
		& R20 (C) & - &  92.44$\pm$0.05 (I, o) & R50 \\  
		& R20 (C) & - &  92.21$\pm$0.06 (P, x) & R50\\
		& R20 (C) & - &  92.25$\pm0$0.04 (P, o) & R50\\ 
		& R20 (C) &  -&  91.9$\pm$0.1 (P+I, x) & R50, R50 \\
		& R20 (C) &  -&  \textbf{92.46$\pm$0.15} (P+I, o) & R50, R50 \\ \hline
\end{tabular}
}
\end{table}

\end{document}